\pdfoutput=1
\documentclass{article}

\usepackage{newtxtext,newtxmath}

\usepackage{hyperref}

\usepackage{amsfonts}       
\usepackage{amsmath}
\usepackage{microtype}      
\usepackage{graphicx}
\usepackage{caption}
\usepackage{subcaption}
\usepackage{natbib}
\usepackage{times}

\usepackage[accepted]{icml2021}

\icmltitlerunning{Emergent Social Learning via Multi-agent Reinforcement Learning}

\begin{document}

\twocolumn[
\icmltitle{Emergent Social Learning via Multi-agent Reinforcement Learning}

\begin{icmlauthorlist}
\icmlauthor{Kamal Ndousse}{oai}
\icmlauthor{Douglas Eck}{goo}
\icmlauthor{Sergey Levine}{goo,bky}
\icmlauthor{Natasha Jaques}{goo,bky}
\end{icmlauthorlist}

\icmlaffiliation{oai}{OpenAI, San Francisco, CA, USA}
\icmlaffiliation{goo}{Google Research -- Brain team, Mountain View, CA, USA}
\icmlaffiliation{bky}{UC Berkeley, Berkeley, CA, USA}

\icmlcorrespondingauthor{Kamal Ndousse}{kamal.ndousse@gmail.com}

\icmlkeywords{Machine Learning, MARL, Multi-Agent Reinforcement Learning, Social Learning, ICML}

\vskip 0.3in
]

\printAffiliationsAndNotice{}

\begin{abstract}
Social learning is a key component of human and animal intelligence. By taking cues from the behavior of experts in their environment, social learners can acquire sophisticated behavior and rapidly adapt to new circumstances. This paper investigates whether independent reinforcement learning (RL) agents in a multi-agent environment can learn to use social learning to improve their performance. We find that in most circumstances, vanilla model-free RL agents do not use social learning. We analyze the reasons for this deficiency, and show that by imposing constraints on the training environment and introducing a model-based auxiliary loss we are able to obtain generalized social learning policies which enable agents to: i) discover complex skills that are not learned from single-agent training, and ii) adapt online to novel environments by taking cues from experts present in the new environment. In contrast, agents trained with model-free RL or imitation learning generalize poorly and do not succeed in the transfer tasks. By mixing multi-agent and solo training, we can obtain agents that use social learning to gain skills that they can deploy when alone, even out-performing agents trained alone from the start.
\end{abstract}

\section{Introduction}
Social learning---learning by observing the behavior of other agents in the same environment---enables both humans and animals to discover useful behaviors that would be difficult to obtain through individual exploration, and to adapt rapidly to new circumstances \citep{Henrich2003,Laland2004socialstrategies}. For example, fish are able to locate safe sources of food in new environments by observing where other members of their species congregate \citep{Laland2004socialstrategies}.
Humans are able to learn from one another without direct access to the experiences or memories of experts. Rather, the interactions that constitute social learning are mediated directly by the environment. 
A novice may gain the ability to solve a task merely by watching a self-interested expert complete the same task, and without an explicit incentive to mimic the expert.
Beyond simple imitation, social learning allows individual humans to ``stand on the shoulders of giants'' and develop sophisticated behaviors or innovations that would be impossible to discover from scratch within one lifetime. In fact, social learning may be a central feature of humans’ unprecedented cognitive and technological development \citep{humphrey1976social, Boyd2011}.

Contemporary techniques for transmitting skills between agents in AI systems are much more limited. For example, Imitation Learning (IL) depends on manually curated expert demonstration trajectories. This requires a system extrinsic to the learning agent ({\em i.e.}, a developer) to (1) identify tasks of interest, (2) find and instrument task experts, and (3) collect demonstration data. The resulting agent is brittle, only able to replicate skills within the collected demonstrations, and otherwise vulnerable to compounding error \cite{ross2011reduction,reddy2019sqil}. In contrast, humans and animals capable of social learning are able to identify experts at tasks relevant to their own interests and learn from those experts without the intervention of an external system. In novel circumstances, social learning enables \textit{adapting online} to the new task, which IL cannot do.

In this work, we investigate whether RL agents can {\em learn} to learn from the cues of other agents. We focus on two hypotheses related to the benefits of social learning: \textbf{H1}) Social learning can discover more \textbf{complex policies} that would be difficult to discover through costly individual exploration, and \textbf{H2}) Social learning can be used to \textbf{adapt online} to a new environment, when the social learning policy encodes how to learn from the cues of experts. We study a setting representative of real-world human environments: multi-agent reinforcement learning (MARL) where completely independent agents pursue their own rewards with no direct incentive to teach or learn from one another (decentralized training and decentralized execution, with no shared parameters or state). This setting is representative of real-world multi-agent problems such as autonomous driving, in which there are other cars on the road but the drivers are not motivated to teach a learning agent, and may or may not have relevant expertise. Novice agents are able to see the experts in their partial observations, but must learn the social learning behavior that would allow them to make use of that information.  In these challenging scenarios, we find that vanilla model-free RL agents fail to use social learning to improve their performance.

Since social learning does not occur by default, we thoroughly investigate 
the environment conditions and agent abilities which are needed for it to emerge, and propose both a training environment and model architecture that facilitate social learning. 

We show that when social learning does occur, it enables agents to learn more complex behavior than agents trained alone and even the experts themselves. 

Within each episode, social learners selectively follow experts only when doing so is rewarding --- which pure imitation learning cannot do.
To prevent social learners from becoming reliant on the presence of experts, we interleave training with experts and training alone. Social learners trained in this manner make use of what they learned from experts to improve their performance in the solo environment, even out-performing agents that were only ever trained solo. This demonstrates that agents can acquire skills through social learning that they could not discover alone, and which are beneficial even when experts are no longer present.

We show for the first time that social learning improves generalization for RL agents.
When evaluated in novel environments with different experts, social learners use cues from the new experts to adapt online, achieving high zero-shot transfer performance. They significantly out-perform agents trained alone, agents that do not learn social learning, and imitation learning. Imitation learning leads to similar transfer performance as the original experts, whereas social learning enables agents to actively seek information about the new environment from a new set of experts and adjust their policy accordingly. The difference between imitation learning and social learning can best be understood using the following analogy: Teach someone to fish, and they will learn how to fish. Teach someone to acquire skills from other agents, and they can learn how to fish, hunt, swim, and any other skills they require.

The contributions of this work are to show
how learning social learning---specifically, learning how to learn from other agents using RL---can lead to improved generalization to novel environments, reduce the cost of individual exploration, and enhance performance. We show how both algorithm and environment design are critical to facilitating social learning, and provide details for how to achieve it. Since deep RL often generalizes poorly 
\citep{cobbe2019quantifying,farebrother2018generalization,packer2018assessing}, we believe these results are a promising first step on a new path for enhancing generalization. In real-world tasks, such as autonomous driving or robotics, there are many human experts who know how to perform useful tasks, but individual exploration can be costly, inefficient, error-prone, or unsafe. Social learning could enable agents to adapt online using cues from these experts, without resorting to clumsy or costly exploration.

\section{Related Work}
There is a rich body of work on imitation learning, which focuses on training an agent to closely approximate the behavior of a curated set of expert trajectories  \citep{pomerleau1989alvinn,schaal1999imitation,billard2008survey,argall2009survey}. For example, Behavioral Cloning (e.g. \citet{bain1995framework,torabi2018behavioral}) uses supervised learning to imitate how the expert maps observations to actions. Because imitation learning can be brittle, it can also be combined with RL (e.g. \citet{guo2019hybrid,lerer2019learning}), or make use of multiple experts (e.g. \citet{cheng2020policy}). Third-person imitation learning (e.g. \citet{shon2006learning,calinon2007learning,stadie2017third}) enables agents to learn to copy an expert's policy when observing the expert from a third-person perspective. 

Our work shares some similarities with third-person imitation learning, in that our agents are never given access to expert's observations and therefore only have a third-person view. However, we do not assume access to a curated set of expert trajectories. Rather, our agents co-exist in a multi-agent environment with other agents with varying degrees of expertise, who are not always within their partially observed field of view. 
We do not assume experts are incentivized to teach other agents (as in e.g. \citet{omidshafiei2019learning,christiano2017deep}). 
Since agents are not given privileged access to information about other agents, they do not explicitly model other agents (as in e.g. \citet{albrecht2018autonomous,jaques2018intrinsic}), but must simply treat them as part of the environment. %
Finally, unlike in imitation learning, we do not force our agents to copy the experts' policy; in fact, we show that our agents eventually learn to out-perform the experts with which they learned. 

Our work is distinct from Inverse RL (IRL) (e.g. \citet{ng2000algorithms,ramachandran2007bayesian,ziebart2008maximum,hadfield2016cooperative}), because our agents share a single environment with the experts which has a consistent reward function, and thus do not need to infer the reward function from expert trajectories.\footnote{Note that novices and experts do not share rewards; if an expert receives a reward the novice does not benefit.} However, we do consider learning from sub-optimal experts, which was studied by \citet{jacq2019learning} in the context of IRL.

Learning online from other agents has been studied in the context of autonomous driving, where \citet{landolfi2018social} trained cars which copied the policy of human drivers on the road when the agent had a high degree of uncertainty. However, they assume access to privileged information about other cars' states and policies. \citet{sun2019behavior} make similar assumptions, but use other cars to update agents' beliefs; e.g. about the presence of pedestrians in occluded regions. 

Most closely related to our work is that of \cite{borsa2019observational}, who train model-free RL agents in the presence of a scripted expert. In the partially observed exploration tasks they consider, they find that that novice agents learning from experts outperform solitary novices only when experts use information that is hidden from the novices. In our work, experts do not have access to privileged information about the environment; rather they are distinguished by their task skill. We analyze why it is difficult for RL agents to benefit from expert demonstrations in sparse reward environments, and propose a method to solve it. With our proposed learning algorithm, novices learning from unprivileged experts do outperform solitary learners. Unlike \citet{borsa2019observational}, we show for the first time that social learning allows agents to transfer effectively to unseen environments with new experts, resulting in better generalization than solo learners.

Finally, our approach uses a model-based auxiliary loss, which predicts the next state given the current state, to better enable agents to learn transition dynamics from trajectories in which they received no reward. The idea of using auxiliary losses in deep RL has been explored in a variety of works (e.g. \citet{ke2019learning,weber2017imagination,shelhamer2016loss,jaderberg2016reinforcement}). Model-based RL (MBRL) has also been applied to the multi-agent context (e.g. \citet{krupnik2020multi}). \citet{hernandez2019agent} use predicting the actions of other agents as an auxiliary task, and show that it improves performance on multi-agent tasks. However, they do not study social learning, transfer or generalization, and assume privileged access to other agents' actions.

\section{Learning Social Learning}

We focus on Multi-Agent Partially Observable Markov Decision Process (MA-POMDP) environments defined by the tuple $\langle \mathcal{S}, \mathcal{A}, \mathcal{T}, \mathcal{R}, \mathcal{I}, N \rangle$, where $N$ is the number of agents. The shared environment state is $s \in \mathcal{S}$. $\mathcal{I}$ is an inspection function that maps $s$ to each agent's partially observed view of the world, $s^k$. At each timestep $t$, each agent $k$ chooses a discrete action $a^k_t \in \mathcal{A}$. Agents act simultaneously and there is no notion of turn-taking. Let $\mathcal{A}^N$ be the joint action space.
The transition function depends on the joint action space: $\mathcal{T}: \mathcal{S} \times \mathcal{A}^N \times \mathcal{S} \rightarrow [0,1]$. 
The reward function 
is the same across agents, but each agent receives its own individual reward $r^k_t = \mathcal{R}(s_t, a^k_t)$. 
Each agent $k$ is attempting to selfishly maximize its own, individual reward by learning a policy $\pi^k$ that optimizes the total expected discounted future reward: $J(\pi^k) = \mathbb{E}_\pi \Big[ \sum_{t=0}^{\infty} \gamma^t \, r^k_{t+1} \, | \, s_0 \Big]$, given starting state $s_0$ and discount factor $\gamma \in [0,1]$. Note that agents are trained independently, \textit{cannot directly observe other agents' observations or actions, and do not share parameters}. To simplify notation, when we discuss the learning objectives for a single novice we forego the superscript notation.

\subsection{Model-Free Sparse Reward Social Learning}
To understand how experts present in the same environment could provide cues that enhance learning---and how model-free RL could fail to benefit from those cues---consider an example environment in which agents must first pick up a key, and use it to pass through a door to reach a goal. In this sparse-reward, hard exploration environment, agents only receive a reward for reaching the goal; the rest of the time, their reward is 0. Assume there is a boundless supply of keys, and after an agent passes through the door it closes --- then other agents cannot modify the environment to make the task easier. Instead, an expert agent can demonstrate a novel state $\tilde{s}$ that is difficult to produce through random exploration: opening the door. Ideally, we would like novice agents to learn from this demonstration by updating their internal representation to model $\tilde{s}$, and eventually learn how to produce this state. 

However, if the novice does not receive any reward as a result of observing $\tilde{s}$, model-free RL receives little benefit from the expert's behavior. Assume that the novice agent is learning policy $\pi_{\theta}$ with a policy-gradient objective on a trajectory including the demonstrated state $\tilde{s}_k, k \in (0,T-1)$:
\begin{align}
   &\nabla_{\theta} J(\theta) = \sum_{t=0}^{T-1} \nabla_{\theta} \log \pi_{\theta}(a_t|s_t)R_t
\end{align}
where $R_t = \sum_{t'=t+1}^T \gamma^{t'-t-1}r_{t'}$ is the total reward received over the course of the trajectory. If the agent receives 0 reward during the episode in which the demonstration occurred (e.g. it does not reach the goal), we can see that $\forall t \in (0, T), R_t = \sum_{t'=t+1}^T \gamma^{t'-t-1}(0) = 0$. Therefore
$\nabla_{\theta} J(\theta) = \sum_{t=0}^{T-1} \nabla_{\theta} \log \pi_{\theta}(a_t|s_t)(0) = 0$,
and the novice receives no gradients from the expert's demonstration which allow it to update its policy. 

Temporal Difference (TD) Learning could temporarily mitigate this issue, but as we show in the Appendix, this ability quickly deteriorates. Consider Q-learning, in which $Q(a,s) = \mathbb{E}_\pi \Big[ \sum_{t=0}^{\infty} \gamma^t \, r_{t+1} \, | \, a, s \Big]$ models the total expected reward from taking action $a$ in state $s$. As the agent continues to receive 0 rewards during training, all $Q$ values will be driven toward zero. Even if the agent observes a useful novel state such as $\tilde{s}$, as $Q(a,s) \rightarrow 0, \forall a \in \mathcal{A}, s \in \mathcal{S}$, the Q-learning objective becomes:
\begin{align}
    Q(\tilde{s}, a) = r + \gamma \max_{a'}Q(s',a') = 0 + \gamma0 = 0
\end{align}
Thus, the objective forces the value of $Q(\tilde{s},a)$ to be zero. In this case, modeling transitions into or out of state $\tilde{s}$ is not required in order to produce an output of zero, since all other Q-values are already zero. In both cases, we can see that before a model-free RL agent receives a reward from the environment, it will have difficulty modeling novel states or transition dynamics. This makes learning from the cues of experts particularly difficult. 

Learning to model the expert's policy $\pi^E(a^E|s^E)$ would likely help the novice improve performance on the task. However, the novice does not have explicit access to the expert's states or actions. From its perspective, the other agent's policy is simply a part of the environment transition dynamics. While the true state transition function $s_{t+1} = \mathcal{T}(s_t, a^N_t, a^E_t)$ depends on both the novice's own action $a^N_t$, and the expert's policy (since $a^E_t = \pi^E(s^E_t$)), the novice is only able to observe $p(s^N_{t+1}|s^N_t, a^N_t)$. Therefore, the novice can only obtain knowledge of the expert's policy through correctly modeling the state transitions it observes. Since, as we have argued above, the novice will struggle to model state transitions in the absence of external reward, it will also have difficulty modeling another agent's policy. 

\subsection{Social Learning with Auxiliary Losses} 
\label{sec:aux_losses}
\begin{figure}[t]
\centering
  \includegraphics[width=\linewidth]{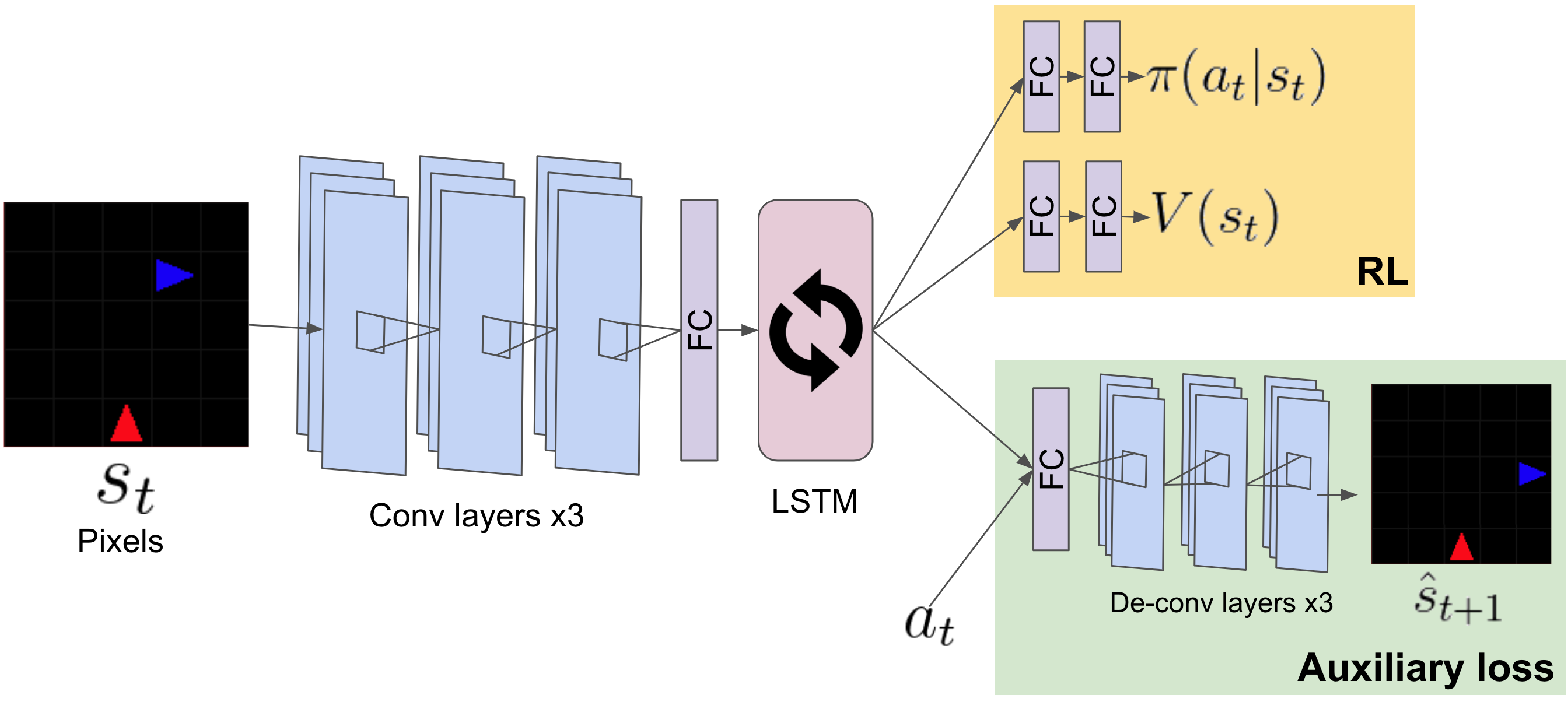}
  \caption{PPO + auxiliary loss deep neural network architecture. Convolution layers extract information about the state from pixels, which is fed into a Fully Connected (FC) layer, and a recurrent LSTM. The yellow shaded box shows the components of the model which learn the RL policy $\pi$ and value function $V(s)$. The green shaded box shows the components of the model dedicated to computing the auxiliary loss, which predicts the next state from the current state.}
  \label{fig:architecture}
  \vspace{-0.3cm}
\end{figure}

\begin{figure*}
  \centering
  \includegraphics[width=\linewidth]{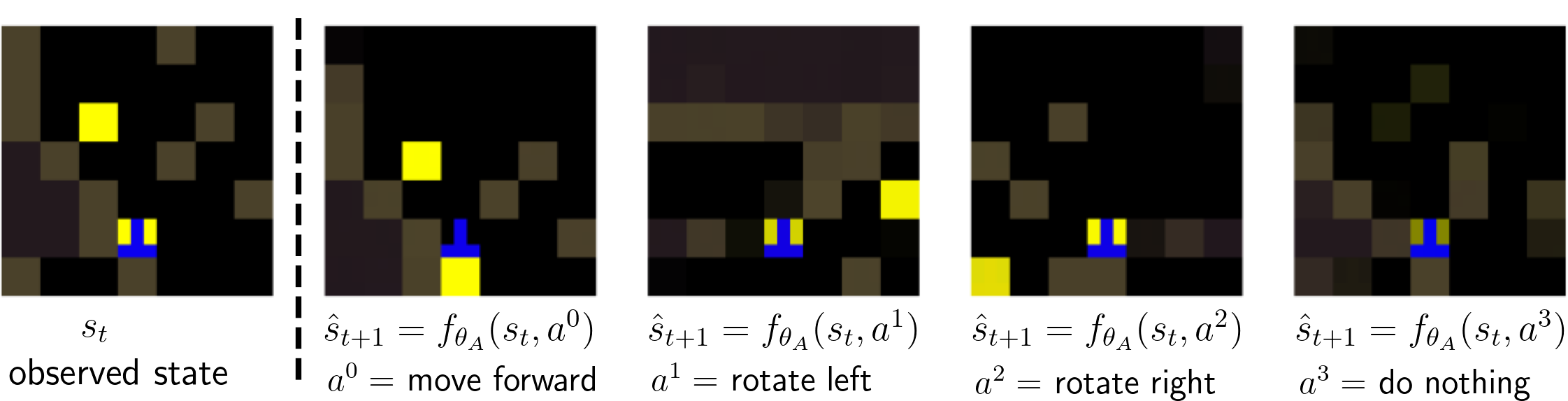}
  \caption{Examples of future states $\hat{s}_{t+1}$ predicted by the network given state $s_t$ (``observed state''), and each of the possible movement actions $a_{t}$. Most predictions are highly accurate, indicating the network has learned to effectively model transition dynamics. The transition for the `do nothing' action is less accurate because it is infrequently chosen by the agent.}
  \label{fig:predictions}
\end{figure*}

To mitigate this issue, we propose augmenting the novice agent with a model-based prediction loss. Specifically, we append additional layers $\theta_A$ to the policy network's encoding of the current state, $f_\theta(s_t)$, as shown in Figure \ref{fig:architecture}. We introduce an unsupervised mean absolute error (MAE) auxiliary loss to train the network to predict the next state $s_{t+1}$, given the current state $s_{t}$ and action $a_{t}$:
\begin{align}
    &\hat{s}_{t+1} = f_{\theta_A}(a_t, s_t); &J = \frac{1}{T}\sum_{t=0}^T |s_{t+1} - \hat{s}_{t+1}|
    \label{eq:mae_next_state}
\end{align}
This architecture allows gradients from the auxiliary loss to contribute to improving $f_\theta(s_t)$. Figure \ref{fig:predictions} shows example state predictions generated by the auxiliary layers of a trained agent, demonstrating that this architecture enables effective learning of the transition dynamics.

We can now see that if the novel demonstration state is in a trajectory, $\tilde{s}_k \in (0,T)$, the term $|\tilde{s}_k - \hat{s}_k|$ will be part of the objective. It will not be 0 unless the agent learns to perfectly predict the novel demonstration state. Therefore, cues from the expert will provide gradients that allow the novice to improve its representation of the world, even if it does not receive any reward from the demonstration. This architecture also implicitly improves the agent's ability to model other agents' policies, since it must correctly predict other agents' actions in order to accurately predict the next state. It is able to do this without ever being given explicit access to the other agent's states or actions, or even any labels indicating what elements of the environment constitute other agents. We refer to our approach as social learning with an auxiliary predictive loss, and hypothesize that it will improve agents' ability to learn from the cues of experts. Note, however, that simply modeling the other agent's policy does not force these independent agents to copy the actions of other agents (unlike in imitation learning). When an independent social learner receives a low reward for following the policy of another agent, it can learn to avoid following that agent in future.

To optimize our agents, we test both TD-learning (deep Q-learning) and policy-gradient methods, and find that Proximal Policy Optimization (PPO) \citep{schulman2017proximal} provides better performance and stability, and is most able to benefit from social learning. We use Generalized Advantage Estimation (GAE) \citep{gae} to train the PPO value function.
As shown in Figure \ref{fig:architecture}, our agents use convolution layers to learn directly from a pixel representation of the state $s_t$. Because the environments under investigation are partially observed and non-Markov, we use a recurrent policy parameterized by a Long Short-Term Memory (LSTM) network \citep{hochreiter1997long} to model the history of observations in each episode. LSTM hidden states are stored in the experience replay buffer. Following the recommendations of \citet{what_matters}, we recalculate the stored state advantage values used to compute the value function loss between mini-batch gradient updates. In addition, we recalculate and update the stored hidden states as suggested by \citet{kapturowski2018recurrent} for off-policy RL. A detailed description of the network architecture and hyperparameters are given in an appendix.

\subsection{Social Learning Environments}
\label{sec:env}

	Humans and animals are most likely to rely on social learning when individual learning is difficult or unsafe \cite{Henrich2003,Laland2004socialstrategies}. Further, individuals prefer to learn from others that they perceive to be highly successful or competent, which is known as \textit{prestige bias} \citep{jimenez2019prestige}. Cues or signals associated with prestige have shown to be important to both human and animal social learning \cite{barkow1975prestige,horner2010prestige}.	
		
\begin{figure*}[t]
	\centering
	\begin{subfigure}{.3\linewidth}
	  \centering
	  \includegraphics[width=.9\linewidth]{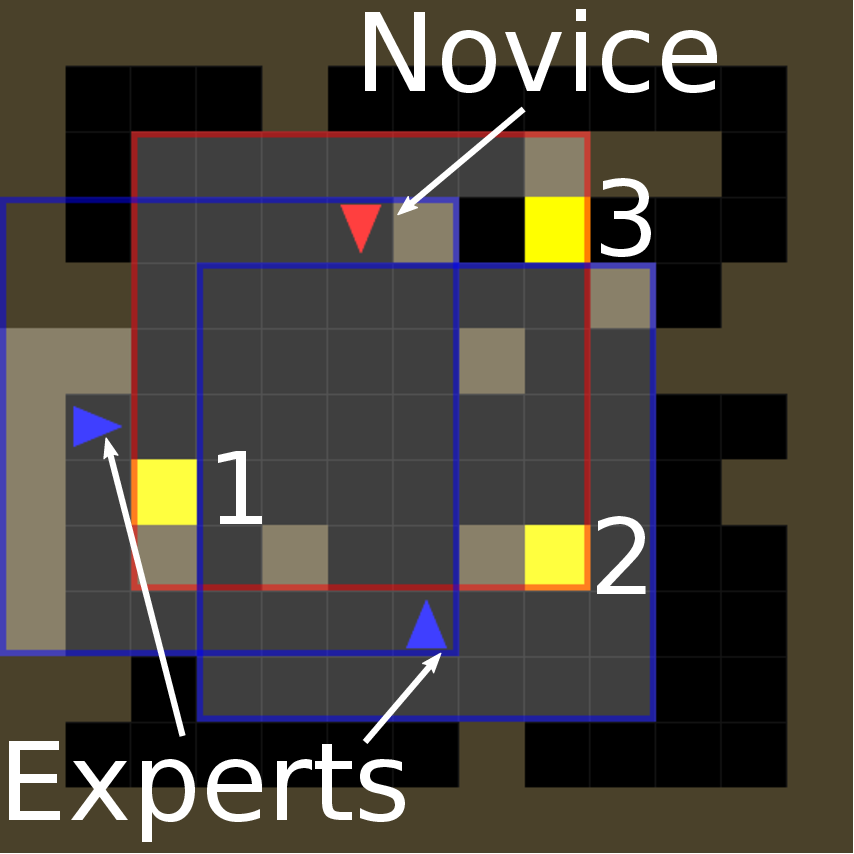}
	  \caption{Goal Cycle}
	  \label{fig:3goal_env}
	\end{subfigure}\hfil%
	\begin{subfigure}{.3\linewidth}
	  \centering
	  \includegraphics[width=.9\linewidth]{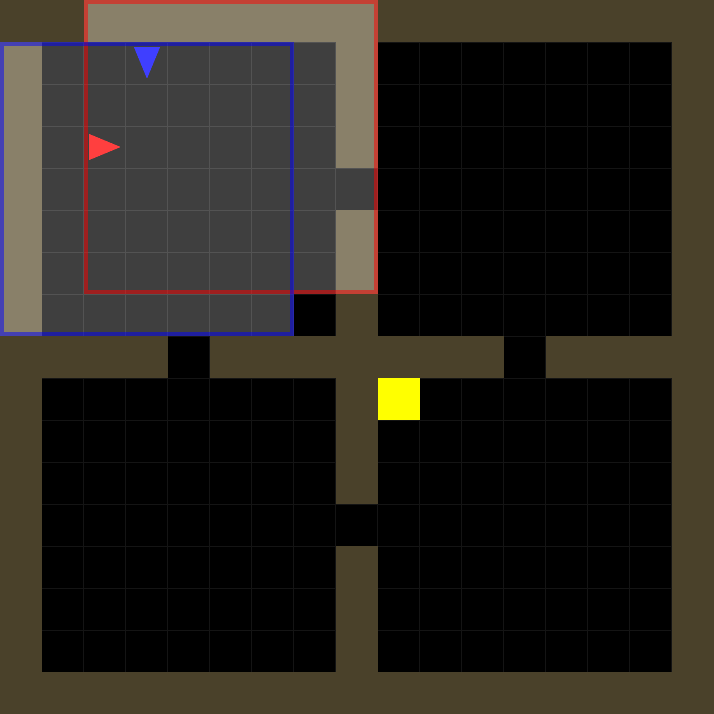}
	  \caption{Four Rooms}
	  \label{fig:4room_env}
	\end{subfigure}
	\begin{subfigure}{.3\linewidth}
	  \centering
	  \includegraphics[width=.9\linewidth]{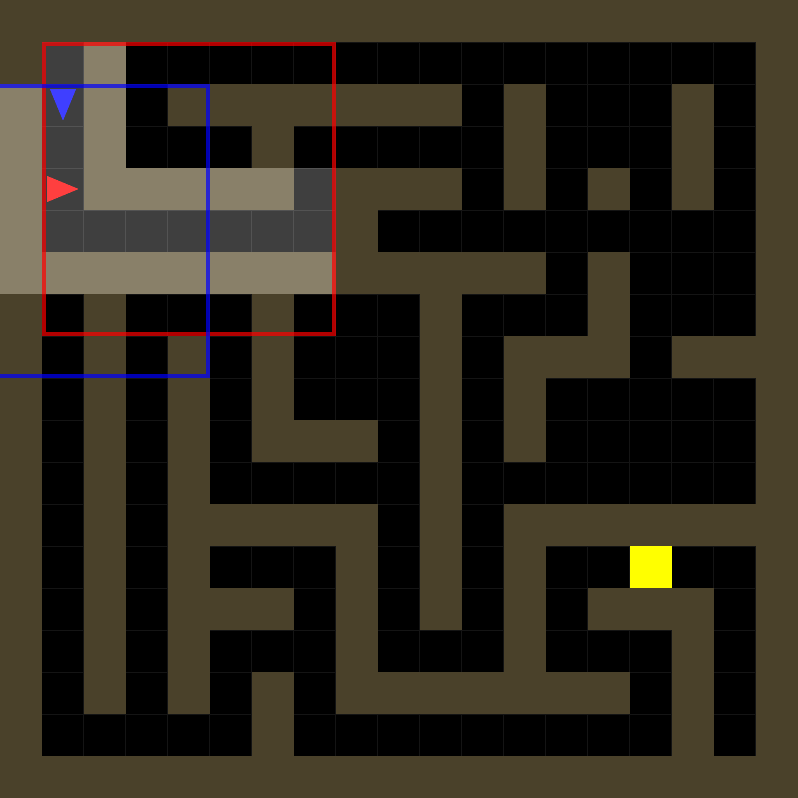}
	  \caption{Maze}
	  \label{fig:maze_env}
	\end{subfigure}
	\caption{Goal Cycle (a) is a 13x13 environment in which the positions of goal tiles (yellow) and obstacles (brown) are randomized at the start of each episode. Agents receive rewards for repeatedly traversing the goal tiles in a certain order 
	and penalties for deviations from the correct order. However, the correct order is not visible. Agents always receive a reward from the first goal tile they encounter in an episode. We test whether agents trained in Goal Cycle can generalize to Four Rooms (b), and Maze (c); larger environments with walls hiding the location of a single goal.}
	\label{fig:goal_cycle}
	\end{figure*}

	Motivated by these two ideas, we introduce a novel environment specifically designed to encourage social learning by making individual exploration difficult and expensive, and introducing prestige cues. The code for both the environment and social learning agents is available at: \url{https://github.com/kandouss/marlgrid}. In Goal Cycle (Figure \ref{fig:3goal_env}), agents are rewarded for navigating between several goal tiles in a certain order, and penalized for deviating from that order. The goal tiles are placed randomly and are visually indistinguishable, so it is not possible for an agent to identify the correct traversal order without potentially incurring an exploration penalty. When the penalty is large, this becomes a hard exploration task.
	
	Agents can achieve high rewards by efficiently traversing between the goals in the correct order, and skilled agents are able to identify the correct order while incurring minimal penalties. Agents can accomplish this either by trial and error, or by observing other agents.
	In practice, since the behavior of other agents can be unpredictable and potentially non-stationary, agents more easily learn to solve the task directly through trial and error.
	But by adjusting the penalty for navigating to goals in the wrong order, we can penalize individual exploration and thereby encourage social learning. 
	
	In all the Goal Cycle variants discussed here, agents receive a reward of $+1$ for navigating to the first goal they encounter in an episode, and $+1$ for any navigating to any subsequent goals if in the correct order. They receive a penalty of $-1.5$ for navigating to the incorrect goal. In Appendix \ref{app:no_penalty} we conduct experiments to assess the effect of the exploration penalty, and find that with no exploration penalty agents do not learn to rely on social learning.

	Prestige cues are implemented in Goal Cycle signal through agents changing color as they collect rewards over the course of each episode. At time $t$, the color of an agent is a linear combination of the red and blue RGB color vectors:
	    $\textsc{color}_{t} = \textsc{blue}\cdot\tilde{c}_{t} + \textsc{red}\cdot(1-\tilde{c}_{t})$,
	where $\tilde{c_{t}} = \text{sigmoid}(c_{t}),$ and $c_{t}$ is a reward-dependent prestige value given by:
	\begin{equation}
	    c_{t} =
	    \begin{cases}
	      \alpha_{c} c_{t-1} + r_{t}, & r_{t} \geq 0 \\
	      0, & \text{otherwise.}
	    \end{cases}
	    \label{eq:prestige_equation}
	\end{equation}
	Thus the color of each agent changes from red to blue as it collects rewards (slowly decaying with constant $\alpha_{c}$, but reverts to red if it incurs a penalty.
	
	The Goal Cycle environments are challenging to master because there are many more incorrect than correct goal traversal sequences. When the penalty is large, the returns for most traversal orders are negative, so mistake-prone agents learn to avoid goal tiles altogether (after the first goal in an episode, which always gives a reward of $+1$). We can obtain agents that perform well in high penalty environments using a curriculum: we train them initially in low-penalty environments and gradually increase the penalty.

\subsection{Social Learning in New Environments}
A central thesis of this work is that social learning can help agents adapt more rapidly to novel environments. This could be particularly useful because deep RL often fails to generalize to even slight modifications of the training environment. Therefore, we test the zero-shot transfer performance of agents pre-trained in 3-Goal Cycle in three new environments. The first is Goal Cycle with 4 goals. This variant is significantly harder, because while there are two possible cycles between three goals, there are six ways to cycle between four goals (there are $(n-1)!$ distinct traversal orders, since the cycle can be entered at any goal). Thus, even optimal agents must incur higher penalties to identify the correct traversal order in each episode.

We also test transfer to the 17x17 Four Rooms (in Figure \ref{fig:4room_env}), and 19x19 Maze environments (Figure \ref{fig:maze_env}). In both environments, the goal and obstacles are randomly generated each episode, and agents must navigate around walls to find the goal within a time limit. These environments represent challenging transfer tasks, because navigating the maze is significantly more complex than navigation in Goal Cycle, and agents are unlikely to have encountered walls in Four Rooms while training in Goal Cycle \cite{dennis2020emergent}.

\begin{figure}[t]
\centering
  \centering
  \includegraphics[width=.9\linewidth]{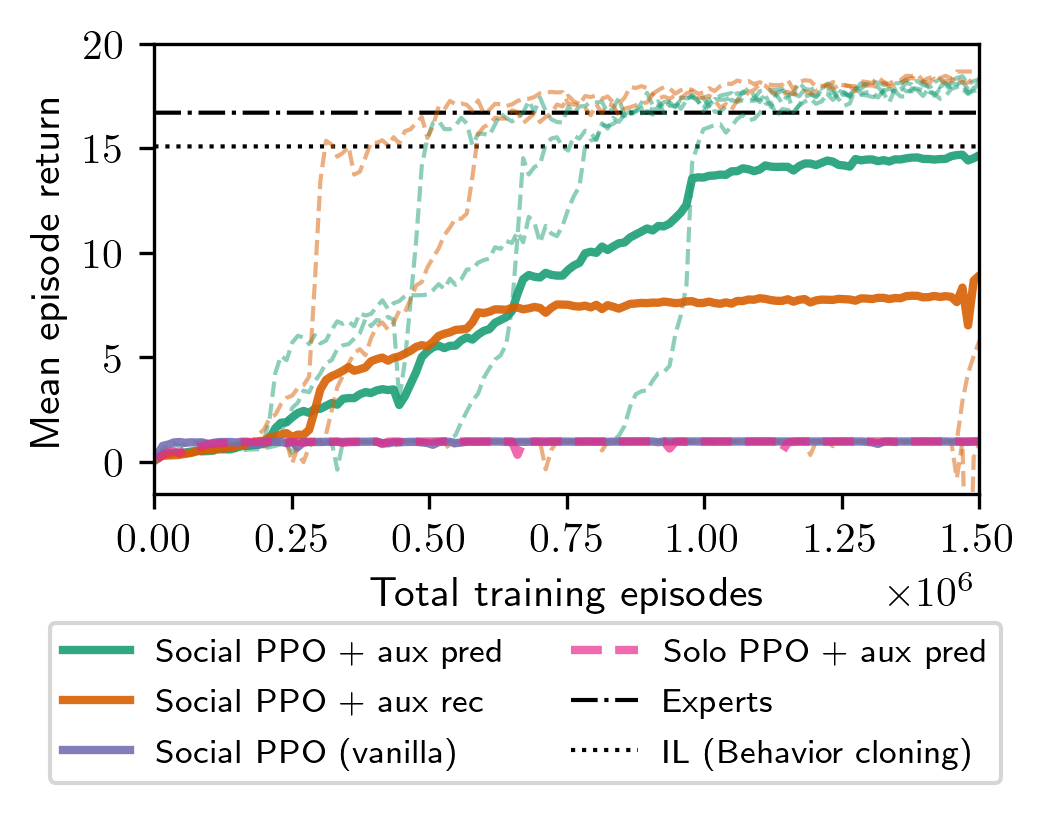}
  
  \caption{While learning in the presence of experts, only agents with an auxiliary loss that enables them to model expert cues (with either a predictive (pred) or reconstructive (rec) auxiliary loss) succeed in using social learning to improve performance.
  None of the seeds for solo agents are able to solve the task. Further, no vanilla PPO seeds solve the task, showing that social learning does not occur automatically with RL. In contrast, 4 of 5 seeds with an auxiliary reconstructive loss (\textsc{social ppo + aux rec}) are able to exceed the performance of the experts present in their environment.  Faded dotted lines show the performance of individual random seeds and bolded lines show the mean of 5 seeds. 
  The final performance is bimodal, with some novice seeds achieving expert-level performance and others failing to learn entirely. Since the normality assumption is violated, we refrain from using confidence intervals.}
  \label{fig:main_result}
  \vspace{-0.3cm}
\end{figure}

\section{Experiments}
\label{sec:experiments}
To establish that independent agents with the auxiliary predictive loss described in Sec.~\ref{sec:aux_losses} can use social learning to improve their own performance, we compare the performance of agents with auxiliary predictive losses learning in an environment shared with experts (\textsc{social ppo + aux pred}) to that of agents with the same architecture but trained alone (Solo PPO + aux pred).

We also compare social learning performance (in the presence of experts) across two ablations. The first uses the same architecture with no auxiliary loss, which corresponds to standard PPO. The second replaces the next-state-prediction objective of Equation \ref{eq:mae_next_state} with a simple autoencoder reconstruction objective that predicts the \textit{current} state, $\hat{s}_{t} = f_{\theta_A}(a_t, s_t)$. The loss is thus $J = \frac{1}{T}\sum_{t=0}^T |s_{t} - \hat{s}_{t}|$. We refer to this as social learning with an auxiliary reconstruction loss (Solo PPO + aux rec).

Unlike the auxiliary prediction loss, this reconstruction loss does not need to learn about transition dynamics or other agents' policies.  Finally, we compare to imitation learning (IL), which is an upper bound on the performance that can be obtained when learning from other agents with direct access to their ground truth state and action information (which is not provided to the other methods).

\subsection{H1: Learning Social Learning}
Figure \ref{fig:main_result} compares Goal Cycle performance of agents trained alone (solo) to agents trained in the presence of experts (social). Solo agents, even with the predictive auxiliary loss (\textsc{solo ppo + aux pred}), are not able to discover the strategy of reaching the goals in the correct order; instead, all random seeds converged to a strategy of stopping after reaching a single goal (with a maximum return of 1). Even when trained in the presence of experts, agents without an auxiliary loss (\textsc{social ppo (vanilla)})

failed in the same way, with all seeds receiving a maximum reward of 1. This supports our hypothesis that model-free RL by itself has difficulty benefiting from the cues of experts in sparse reward environments. 

\begin{figure}[t]
\centering
  \centering
  \includegraphics[width=.9\linewidth]{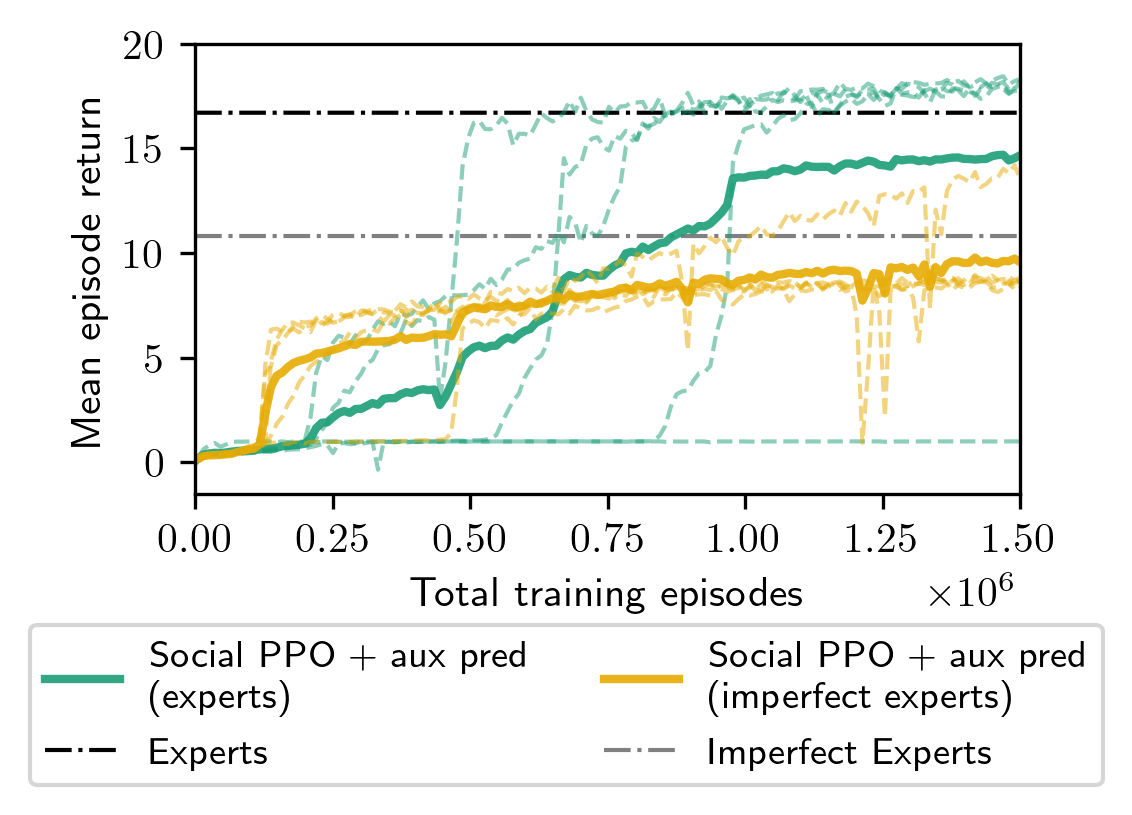}

  \caption{The effect of expert skill on social learning. With auxiliary predictive losses, novice agents benefit from social learning in the presence of near-optimal experts (green) or imperfect experts (yellow), surpassing the performance of solo agents in both cases. However, agents learning from imperfect experts achieve lower performance, and only 1/5 random seeds exceed the performance of the imperfect experts.}
  \label{fig:imperfect}
  \vspace{-0.5cm}
\end{figure}

In contrast, social agents trained with auxiliary predictive or reconstructive losses were able to achieve higher performance. This confirms hypothesis \textbf{H1}, that when successful, social learning can enable agents to learn more complex behavior than when learning alone. While the reconstruction loss helps agents learn from experts, still higher performance is obtained with the model-based prediction loss. In contrast to the reconstruction loss, the prediction loss helps agents learn representations that capture the transition dynamics and (implicitly) other agents` policies. Note that in our experiments, solo agents with auxiliary predictive losses performed much worse than social learners, showing that good performance can depend on learning effectively from expert cues in sparse reward hard-exploration environments. 

\begin{figure*}[t]
\centering
\begin{subfigure}{0.33\linewidth}
  \centering
  \includegraphics[width=\linewidth]{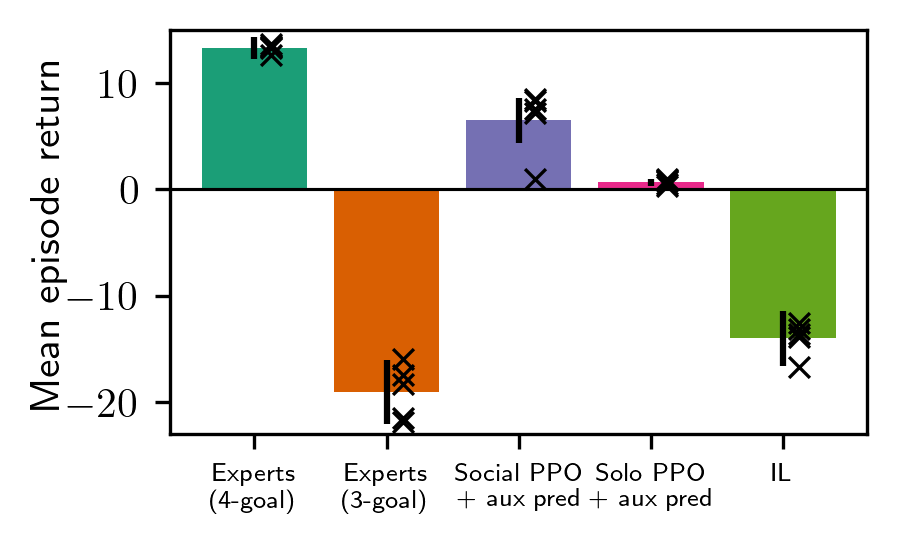}

  \caption{Goal Cycle (4 goals, with experts)}
  \label{fig:4goal}
\end{subfigure}\hfil%
\begin{subfigure}{0.33\linewidth}
  \centering
  \includegraphics[width=\linewidth]{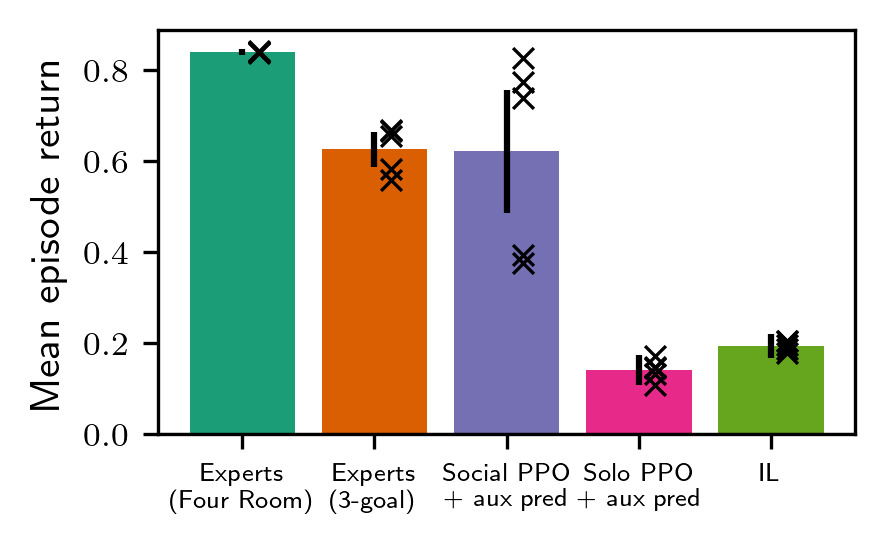}
  \caption{Four Rooms (with expert)}
  \label{fig:4room}
\end{subfigure}
\begin{subfigure}{0.33\linewidth}
  \centering
  \includegraphics[width=\linewidth]{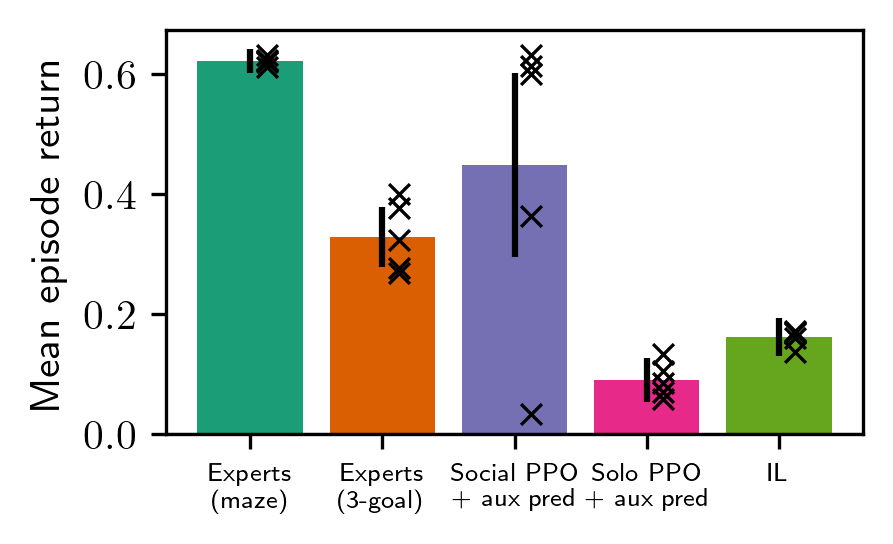}
  \caption{Maze (with expert)}
  \label{fig:mazetransfer}
\end{subfigure}
\caption{Zero-shot transfer performance in novel environments with expert demonstrators. Each bar shows the mean of the multiple seeds, with the performance of each seed shown as a black `x'. 
The experts and 3-goal solo learners were also trained with auxiliary predictive losses. Error bars show $95\%$ confidence intervals for the estimated means. 
Agents which have learned to rely on social learning benefit from the cues of novel experts in the new environment., and easily out-perform agents trained alone, agents training with imitation learning, and in some cases, the experts from which they originally learned. Videos: tinyurl.com/SociAPL}
\label{fig:transfer}
\end{figure*}

Interestingly, the majority of random seeds with predictive losses are actually able to exceed the performance of both the experts 
and the imitation learning (IL) methods. Unlike IL, which requires direct access to experts' states and actions, social learners reach equivalent performance only through learning from their partially observed view of experts via RL. The reason that social learners can actually exceed the performance of the experts themselves is that they learn an improved policy that leverages social information. During each episode, social learners wait and observe while the expert explores the goals to find the correct order. Only then do they follow the expert's cues. Thus, they allow the expert to incur the exploration penalties of Goal Cycle and use social learning to follow when it is safe. Despite being a non-cooperative strategy, this is reminiscent of social learning in the animal kingdom \cite{laland2018darwin}.

\textbf{Learning from Sub-Optimal Experts.}
In real-world environments, not all other agents will have optimal performance. Therefore, we test whether it is still possible to learn from imperfect experts. The results in Figure \ref{fig:imperfect} show that while expert performance does affect the novices' final performance, even when learning from imperfect experts novices can use social learning to exceed the performance of agents trained alone. However, novices trained with optimal experts more frequently learn to exceed the performance of the experts with which they are trained.

\subsection{H2: Adapting Online to New Environments}

We hypothesize that using social learning to adapt online within an episode will allow agents to achieve higher performance when transferred to a new environment with experts. To evaluate this hypothesis, we
investigate how well agents pre-trained in 3-Goal Cycle can generalize to three new environments: a 4 goal variant of the Goal Cycle environment, Four Rooms, and Maze. Figure~\ref{fig:transfer} shows the zero-shot transfer performance of agents trained in 3-Goal Cycle to each new environment. Agents trained alone (which have not learned to rely on social learning) perform poorly, likely because they were never able to obtain good exploration policies during training, due to the high cost of exploration in the training environment. 

Imitation learning agents perform poorly across all three transfer environments, under-performing the original 3-Goal experts in every case. These results demonstrate the brittleness of imitation learning, which can only fit the expert trajectories, and generalizes poorly to states outside of the training distribution \cite{ross2011reduction,reddy2019sqil}. In contrast, agents trained with social PPO + aux pred loss employ a generalized social learning policy that encodes how to adapt online based on expert cues. Thus, these agents use social learning to achieve high performance in the new environments with a new set of experts. For example, in the Maze environment they can follow the cues of the Maze experts to more closely approximate the optimal policy, rather than continuing to rely on learning from 3-Goal experts. 

It is worth noting that the original 3-Goal experts 
transfer poorly to both 4-Goal Cycle and Maze. In 4-Goal, they receive large negative scores, repeatedly incurring a penalty for navigating to the goals in the incorrect order. This illustrates the brittleness of an RL policy which is overfit to the training environment. In both environments, social learning agents achieve better performance on the transfer task than the original experts, suggesting social learning may be a generally beneficial strategy for improving transfer in RL. Taken together, these results provide support for hypothesis \textbf{H2}, and suggest social learning can enable RL agents to adapt online (within an episode) to new environments by taking cues from other agents.

\begin{figure*}[t]
\centering
\begin{subfigure}{0.33\linewidth}
  \centering
  \includegraphics[width=\linewidth]{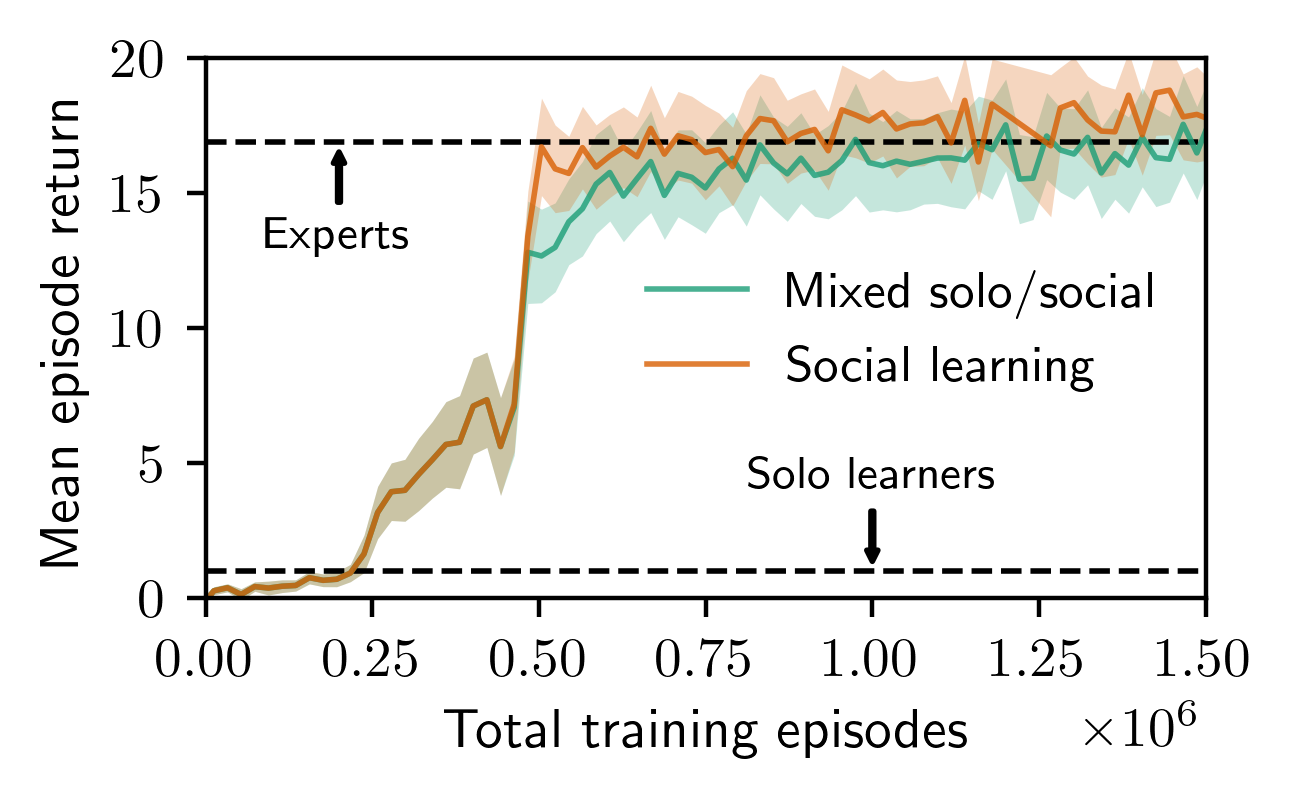}
  \caption{Training with experts}
  \label{fig:cake_3goal_social}
\end{subfigure}\hfil%
\begin{subfigure}{0.33\linewidth}
  \centering
  \includegraphics[width=\linewidth]{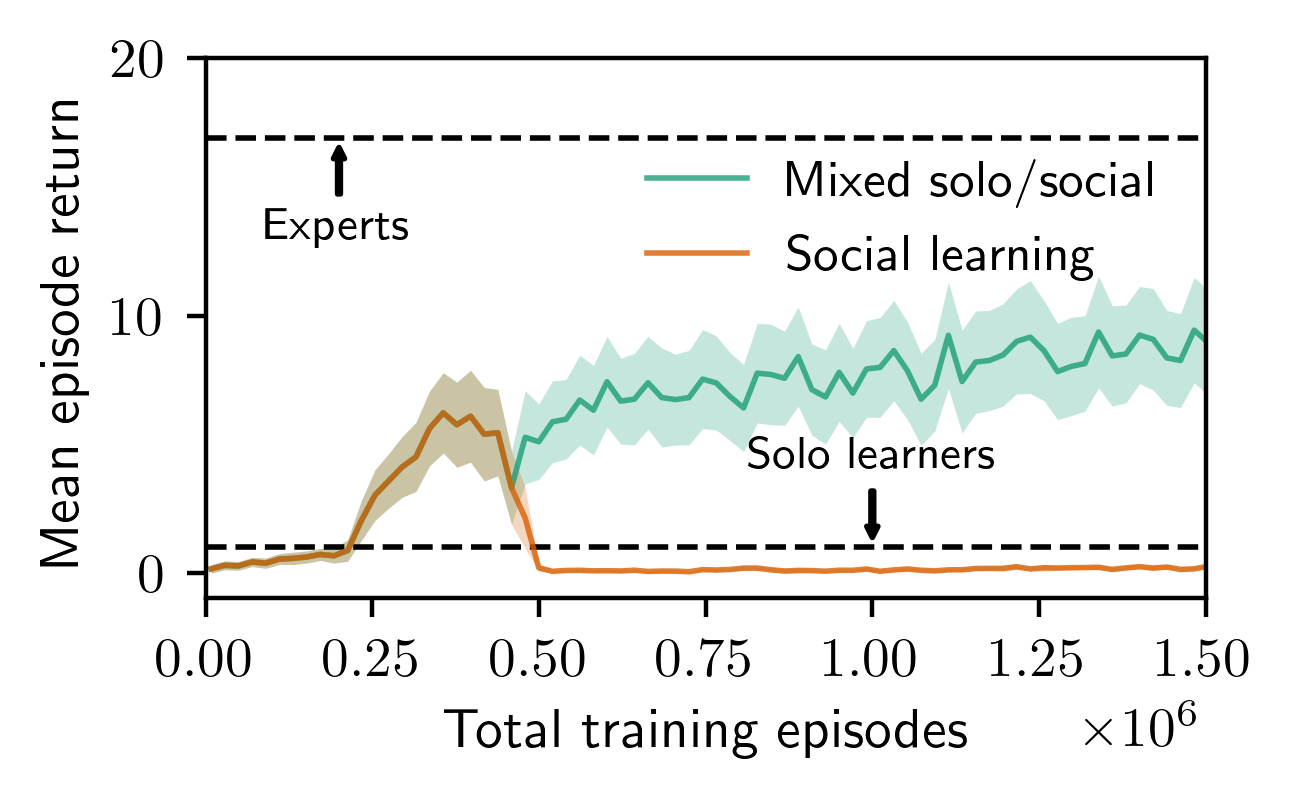}
  \caption{Transfer to solo 3-Goal}
  \label{fig:cake_3goal_solo}
\end{subfigure}
\begin{subfigure}{0.33\linewidth}
  \centering
  \includegraphics[width=\linewidth]{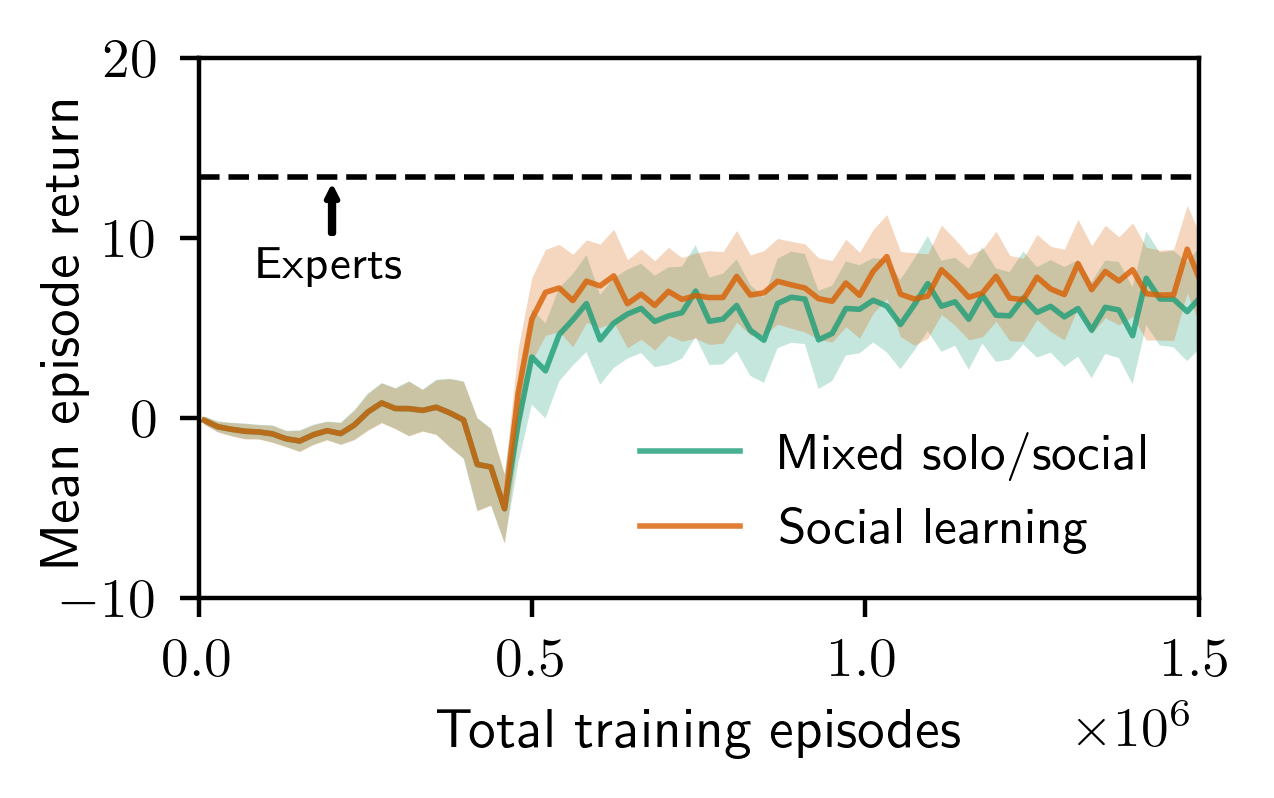}
  \caption{Transfer to 4-Goal with experts}
  \label{fig:cake_4goal_social}
\end{subfigure}
\caption{ \textbf{\textsc{Orange~lines:}} performance throughout training for one of the \textsc{social ppo + aux pred} seeds from Figure \ref{fig:main_result} evaluated in three Goal Cycle environments: (a) the training environment, 3-goal Goal Cycle with experts; (b) a solo version of the training environment (no experts); and (c) 4-Goal Cycle, with experts. The agent initially relies mostly on individual exploration as indicated by good zero-shot transfer performance in the solo environment. After about 500k episodes the agent becomes reliant on cues from the 3-Goal experts, and its performance in solo 3-goal Goal Cycle degrades. \hspace{10pt}~\textbf{\textsc{Green~lines:}}
performance of the same agent if the distribution of training environments is changed from 100\% social to 75\% solo and 25\% social after about 500k episodes. As training proceeds the agent retains the capacity to solve the solo 3-goal environment while learning to use cues from expert behavior when they are available. The performance of this agent in the solo environment actually exceeds that of agents trained exclusively in the solo environment. The agent retains the capacity to use social learning when possible, with performance in the social training environment (a) and 4-goal Goal Cycle with experts (c) comparable to the orange agent (for which 100\% of training episodes included experts).
}
\label{fig:cake}
\end{figure*}

\subsection{Social Learning Helps Solo Learning}

A potential downside to learning to rely on the cues of experts is that agents could fail to perform well when experts are removed from the environment. For novice agents always trained with experts, there appears to be a trade-off between social learning and individual learning. As shown in Figure~\ref{fig:cake}, novice \textsc{social ppo + aux pred} agents initially learn to solve the Goal Cycle task with individual exploration, but eventually they overfit to the presence of experts and become reliant on expert cues to the detriment of solo performance. We can see this because the performance in the 3-goal environment in the presence of experts increases (\ref{fig:cake_3goal_social}), while the performance in the 3-goal environment when alone drops (\ref{fig:cake_3goal_solo}). Observing the agent's behavior reveals that it has learned to follow the cues of experts when they are present in the environment, but refrain from individual exploration when experts are not present. Given the high cost of individual exploration in this environment, this a safe but conservative strategy. 

We find that we can improve the solo performance of social learning agents by changing the distribution of training tasks to include episodes in which the experts are not present. The green curves in Figure \ref{fig:cake} show the performance of a checkpoint switched from entirely social training environments to a mix of solo and social environments as training proceeds. The agent is able to retain good performance in solo 3-goal environments as well as 4-goal environments with experts, indicating that it is learning to opportunistically take advantage of expert cues while building individual expertise in the 3-goal environment. In fact, agents trained with \textsc{social ppo + aux pred} perform better in solo transfer environments than agents exclusively trained in the solo environment (see Figure \ref{fig:cake_3goal_solo}), because agents trained alone never discover the optimal strategy due to the difficulty of exploring in this environment. 
This demonstrates that not only does social learning enable agents to discover skills that they could not learn by themselves (again supporting hypothesis \textbf{H1}), but that they are able to retain this knowledge to improve their individual performance even when experts are not present. 

\section{Conclusions}
In this work we investigated how model-free deep RL can benefit from the presence of expert agents in the environment. We design an environment to elicit social learning by increasing the cost of individual exploration and introducing prestige cues. However, we find that even in this environment, model-free agents fail to use social learning to reach optimal performance. By adding a model-based auxiliary loss, which requires implicitly modeling other agents' policies, we are able to train agents to use social learning. When deployed to novel environments, these agents use their social learning policies to adapt online using the cues of novel experts, and perform well in zero-shot transfer tasks. In contrast, agents trained alone or with imitation learning cannot adapt to the new task. Further, by mixing social and solo training, we obtain social learning agents that actually have higher performance in the solo environment than agents trained exclusively in the solo environment.
Our results demonstrate that social learning not only enables agents to learn complex behavior that they do not discover when trained alone, but that it can allow agents to generalize better when transferred to novel environments.

\subsection{Limitations and Future Work}
Our social learning experiments focus on exploratory navigation tasks. 
Future work includes extending this to other domains such as manipulation, to scenarios in which experts pursue different goals than the novices, and to scenarios with multiple experts that employ a variety of strategies. We found that training in a mixture of social and solitary environments can permit novice social learning agents to develop effective strategies for both social and individual task variants, and notably that the resulting individual skill far exceeds that of a solitary novice. However, in this work we do not thoroughly explore different strategies for augmenting solitary with social experience. Further research could clarify the circumstances in which adding social experiences could aid solitary task performance, and see the development of algorithms to facilitate this for arbitrary tasks. Finally, we assume that prestige cues which indicate how well experts perform in the environment are visible to social learners. Future research could investigate how to accurately predict another agent's performance when prestige cues are not available. 

\section{Acknowledgements}
We are grateful to the OpenAI Scholars program for facilitating the collaboration that led to this paper. We would also like to thank Ken Stanley, Joel Lehman, Bowen Baker, Yi Wu, Aleksandra Faust, Max Kleiman-Weiner, and DJ Strouse for helpful discussions related to this work, and Alethea Power for providing compute resources that facilitated our experiments.

\bibliographystyle{icml2021}
\bibliography{social_learning}

\begin{thebibliography}{53}
\providecommand{\natexlab}[1]{#1}
\providecommand{\url}[1]{\texttt{#1}}
\expandafter\ifx\csname urlstyle\endcsname\relax
  \providecommand{\doi}[1]{doi: #1}\else
  \providecommand{\doi}{doi: \begingroup \urlstyle{rm}\Url}\fi

\bibitem[Albrecht \& Stone(2018)Albrecht and Stone]{albrecht2018autonomous}
Albrecht, S.~V. and Stone, P.
\newblock Autonomous agents modelling other agents: A comprehensive survey and
  open problems.
\newblock \emph{Artificial Intelligence}, 258:\penalty0 66--95, 2018.

\bibitem[Andrychowicz et~al.(2020)Andrychowicz, Raichuk, Stańczyk, Orsini,
  Girgin, Marinier, Hussenot, Geist, Pietquin, Michalski, Gelly, and
  Bachem]{what_matters}
Andrychowicz, M., Raichuk, A., Stańczyk, P., Orsini, M., Girgin, S., Marinier,
  R., Hussenot, L., Geist, M., Pietquin, O., Michalski, M., Gelly, S., and
  Bachem, O.
\newblock What matters in on-policy reinforcement learning? a large-scale
  empirical study.
\newblock \emph{arXiv preprint arXiv:2006.05990}, 2020.

\bibitem[Argall et~al.(2009)Argall, Chernova, Veloso, and
  Browning]{argall2009survey}
Argall, B.~D., Chernova, S., Veloso, M., and Browning, B.
\newblock A survey of robot learning from demonstration.
\newblock \emph{Robotics and autonomous systems}, 57\penalty0 (5):\penalty0
  469--483, 2009.

\bibitem[Bain \& Sammut(1995)Bain and Sammut]{bain1995framework}
Bain, M. and Sammut, C.
\newblock A framework for behavioural cloning.
\newblock In \emph{Machine Intelligence 15}, pp.\  103--129, 1995.

\bibitem[Barkow et~al.(1975)Barkow, Akiwowo, Barua, Chance, Chapple,
  Chattopadhyay, Freedman, Geddes, Goswami, Isichei,
  et~al.]{barkow1975prestige}
Barkow, J.~H., Akiwowo, A.~A., Barua, T.~K., Chance, M., Chapple, E.~D.,
  Chattopadhyay, G.~P., Freedman, D.~G., Geddes, W., Goswami, B., Isichei, P.,
  et~al.
\newblock Prestige and culture: a biosocial interpretation [and comments and
  replies].
\newblock \emph{Current Anthropology}, 16\penalty0 (4):\penalty0 553--572,
  1975.

\bibitem[Biewald(2020)]{wandb}
Biewald, L.
\newblock Experiment tracking with weights and biases, 2020.
\newblock URL \url{https://www.wandb.com/}.
\newblock Software available from wandb.com.

\bibitem[Billard et~al.(2008)Billard, Calinon, Dillmann, and
  Schaal]{billard2008survey}
Billard, A., Calinon, S., Dillmann, R., and Schaal, S.
\newblock Survey: Robot programming by demonstration.
\newblock \emph{Handbook of robotics}, 59\penalty0 (BOOK\_CHAP), 2008.

\bibitem[Borsa et~al.(2019)Borsa, Heess, Piot, Liu, Hasenclever, Munos, and
  Pietquin]{borsa2019observational}
Borsa, D., Heess, N., Piot, B., Liu, S., Hasenclever, L., Munos, R., and
  Pietquin, O.
\newblock Observational learning by reinforcement learning.
\newblock In \emph{Proceedings of the 18th International Conference on
  Autonomous Agents and MultiAgent Systems}, pp.\  1117--1124. International
  Foundation for Autonomous Agents and Multiagent Systems, 2019.

\bibitem[Boyd et~al.(2011)Boyd, Richerson, and Henrich]{Boyd2011}
Boyd, R., Richerson, P.~J., and Henrich, J.
\newblock The cultural niche: Why social learning is essential for human
  adaptation.
\newblock \emph{Proceedings of the National Academy of Sciences}, 108\penalty0
  (Supplement{\_}2):\penalty0 10918--10925, June 2011.

\bibitem[Calinon et~al.(2007)Calinon, Guenter, and
  Billard]{calinon2007learning}
Calinon, S., Guenter, F., and Billard, A.
\newblock On learning, representing, and generalizing a task in a humanoid
  robot.
\newblock \emph{IEEE Transactions on Systems, Man, and Cybernetics, Part B
  (Cybernetics)}, 37\penalty0 (2):\penalty0 286--298, 2007.

\bibitem[Cheng et~al.(2020)Cheng, Kolobov, and Agarwal]{cheng2020policy}
Cheng, C.-A., Kolobov, A., and Agarwal, A.
\newblock Policy improvement from multiple experts.
\newblock \emph{arXiv preprint arXiv:2007.00795}, 2020.

\bibitem[Chevalier-Boisvert et~al.(2018)Chevalier-Boisvert, Willems, and
  Pal]{gym_minigrid}
Chevalier-Boisvert, M., Willems, L., and Pal, S.
\newblock Minimalistic gridworld environment for openai gym.
\newblock \url{https://github.com/maximecb/gym-minigrid}, 2018.

\bibitem[Christiano et~al.(2017)Christiano, Leike, Brown, Martic, Legg, and
  Amodei]{christiano2017deep}
Christiano, P.~F., Leike, J., Brown, T., Martic, M., Legg, S., and Amodei, D.
\newblock Deep reinforcement learning from human preferences.
\newblock In \emph{Advances in Neural Information Processing Systems}, pp.\
  4299--4307, 2017.

\bibitem[Cobbe et~al.(2019)Cobbe, Klimov, Hesse, Kim, and
  Schulman]{cobbe2019quantifying}
Cobbe, K., Klimov, O., Hesse, C., Kim, T., and Schulman, J.
\newblock Quantifying generalization in reinforcement learning.
\newblock In \emph{International Conference on Machine Learning}, pp.\
  1282--1289, 2019.

\bibitem[Dennis et~al.(2020)Dennis, Jaques, Vinitsky, Bayen, Russell, Critch,
  and Levine]{dennis2020emergent}
Dennis, M., Jaques, N., Vinitsky, E., Bayen, A., Russell, S., Critch, A., and
  Levine, S.
\newblock Emergent complexity and zero-shot transfer via unsupervised
  environment design.
\newblock \emph{arXiv preprint arXiv:2012.02096}, 2020.

\bibitem[Farebrother et~al.(2018)Farebrother, Machado, and
  Bowling]{farebrother2018generalization}
Farebrother, J., Machado, M.~C., and Bowling, M.
\newblock Generalization and regularization in dqn.
\newblock \emph{arXiv preprint arXiv:1810.00123}, 2018.

\bibitem[Guo et~al.(2019)Guo, Chang, Yu, Tesauro, and Campbell]{guo2019hybrid}
Guo, X., Chang, S., Yu, M., Tesauro, G., and Campbell, M.
\newblock Hybrid reinforcement learning with expert state sequences.
\newblock In \emph{Proceedings of the AAAI Conference on Artificial
  Intelligence}, volume~33, pp.\  3739--3746, 2019.

\bibitem[Hadfield-Menell et~al.(2016)Hadfield-Menell, Russell, Abbeel, and
  Dragan]{hadfield2016cooperative}
Hadfield-Menell, D., Russell, S.~J., Abbeel, P., and Dragan, A.
\newblock Cooperative inverse reinforcement learning.
\newblock In \emph{Advances in neural information processing systems}, pp.\
  3909--3917, 2016.

\bibitem[Henrich \& McElreath(2003)Henrich and McElreath]{Henrich2003}
Henrich, J. and McElreath, R.
\newblock The evolution of cultural evolution.
\newblock \emph{Evolutionary Anthropology: Issues, News, and Reviews},
  12\penalty0 (3):\penalty0 123--135, May 2003.

\bibitem[Hernandez-Leal et~al.(2019)Hernandez-Leal, Kartal, and
  Taylor]{hernandez2019agent}
Hernandez-Leal, P., Kartal, B., and Taylor, M.~E.
\newblock Agent modeling as auxiliary task for deep reinforcement learning.
\newblock In \emph{Proceedings of the AAAI Conference on Artificial
  Intelligence and Interactive Digital Entertainment}, volume~15, pp.\  31--37,
  2019.

\bibitem[Hochreiter \& Schmidhuber(1997)Hochreiter and
  Schmidhuber]{hochreiter1997long}
Hochreiter, S. and Schmidhuber, J.
\newblock Long short-term memory.
\newblock \emph{Neural computation}, 9\penalty0 (8):\penalty0 1735--1780, 1997.

\bibitem[Horner et~al.(2010)Horner, Proctor, Bonnie, Whiten, and
  de~Waal]{horner2010prestige}
Horner, V., Proctor, D., Bonnie, K.~E., Whiten, A., and de~Waal, F.~B.
\newblock Prestige affects cultural learning in chimpanzees.
\newblock \emph{PloS one}, 5\penalty0 (5):\penalty0 e10625, 2010.

\bibitem[Humphrey(1976)]{humphrey1976social}
Humphrey, N.~K.
\newblock The social function of intellect.
\newblock In \emph{Growing points in ethology}, pp.\  303--317. Cambridge
  University Press, 1976.

\bibitem[Jacq et~al.(2019)Jacq, Geist, Paiva, and Pietquin]{jacq2019learning}
Jacq, A., Geist, M., Paiva, A., and Pietquin, O.
\newblock Learning from a learner.
\newblock In \emph{International Conference on Machine Learning}, pp.\
  2990--2999, 2019.

\bibitem[Jaderberg et~al.(2016)Jaderberg, Mnih, Czarnecki, Schaul, Leibo,
  Silver, and Kavukcuoglu]{jaderberg2016reinforcement}
Jaderberg, M., Mnih, V., Czarnecki, W.~M., Schaul, T., Leibo, J.~Z., Silver,
  D., and Kavukcuoglu, K.
\newblock Reinforcement learning with unsupervised auxiliary tasks.
\newblock \emph{arXiv preprint arXiv:1611.05397}, 2016.

\bibitem[Jaques et~al.(2018)Jaques, Lazaridou, Hughes, Gulcehre, Ortega,
  Strouse, Leibo, and de~Freitas]{jaques2018intrinsic}
Jaques, N., Lazaridou, A., Hughes, E., Gulcehre, C., Ortega, P.~A., Strouse,
  D., Leibo, J.~Z., and de~Freitas, N.
\newblock Intrinsic social motivation via causal influence in multi-agent rl.
\newblock \emph{arXiv preprint arXiv:1810.08647}, 2018.

\bibitem[Jim{\'e}nez \& Mesoudi(2019)Jim{\'e}nez and
  Mesoudi]{jimenez2019prestige}
Jim{\'e}nez, {\'A}.~V. and Mesoudi, A.
\newblock Prestige-biased social learning: current evidence and outstanding
  questions.
\newblock \emph{Palgrave Communications}, 5\penalty0 (1):\penalty0 1--12, 2019.

\bibitem[Kapturowski et~al.(2018)Kapturowski, Ostrovski, Quan, Munos, and
  Dabney]{kapturowski2018recurrent}
Kapturowski, S., Ostrovski, G., Quan, J., Munos, R., and Dabney, W.
\newblock Recurrent experience replay in distributed reinforcement learning.
\newblock In \emph{International conference on learning representations}, 2018.

\bibitem[Ke et~al.(2019)Ke, Singh, Touati, Goyal, Bengio, Parikh, and
  Batra]{ke2019learning}
Ke, N.~R., Singh, A., Touati, A., Goyal, A., Bengio, Y., Parikh, D., and Batra,
  D.
\newblock Learning dynamics model in reinforcement learning by incorporating
  the long term future.
\newblock \emph{arXiv preprint arXiv:1903.01599}, 2019.

\bibitem[Kingma \& Ba(2014)Kingma and Ba]{adam}
Kingma, D. and Ba, J.
\newblock Adam: A method for stochastic optimization.
\newblock \emph{International Conference on Learning Representations}, 12 2014.

\bibitem[Krupnik et~al.(2020)Krupnik, Mordatch, and Tamar]{krupnik2020multi}
Krupnik, O., Mordatch, I., and Tamar, A.
\newblock Multi-agent reinforcement learning with multi-step generative models.
\newblock In \emph{Conference on Robot Learning}, pp.\  776--790, 2020.

\bibitem[Laland(2004)]{Laland2004socialstrategies}
Laland, K.~N.
\newblock Social learning strategies.
\newblock \emph{Animal Learning {\&} Behavior}, 32\penalty0 (1):\penalty0
  4--14, February 2004.

\bibitem[Laland(2018)]{laland2018darwin}
Laland, K.~N.
\newblock \emph{Darwin's unfinished symphony: How culture made the human mind}.
\newblock Princeton University Press, 2018.

\bibitem[Landolfi \& Dragan(2018)Landolfi and Dragan]{landolfi2018social}
Landolfi, N.~C. and Dragan, A.~D.
\newblock Social cohesion in autonomous driving.
\newblock In \emph{2018 IEEE/RSJ International Conference on Intelligent Robots
  and Systems (IROS)}, pp.\  8118--8125. IEEE, 2018.

\bibitem[Lerer \& Peysakhovich(2019)Lerer and Peysakhovich]{lerer2019learning}
Lerer, A. and Peysakhovich, A.
\newblock Learning existing social conventions via observationally augmented
  self-play.
\newblock In \emph{Proceedings of the 2019 AAAI/ACM Conference on AI, Ethics,
  and Society}, pp.\  107--114, 2019.

\bibitem[Ng et~al.(2000)Ng, Russell, et~al.]{ng2000algorithms}
Ng, A.~Y., Russell, S.~J., et~al.
\newblock Algorithms for inverse reinforcement learning.
\newblock In \emph{Icml}, volume~1, pp.\ ~2, 2000.

\bibitem[Omidshafiei et~al.(2019)Omidshafiei, Kim, Liu, Tesauro, Riemer, Amato,
  Campbell, and How]{omidshafiei2019learning}
Omidshafiei, S., Kim, D.-K., Liu, M., Tesauro, G., Riemer, M., Amato, C.,
  Campbell, M., and How, J.~P.
\newblock Learning to teach in cooperative multiagent reinforcement learning.
\newblock In \emph{Proceedings of the AAAI Conference on Artificial
  Intelligence}, volume~33, pp.\  6128--6136, 2019.

\bibitem[Packer et~al.(2018)Packer, Gao, Kos, Kr{\"a}henb{\"u}hl, Koltun, and
  Song]{packer2018assessing}
Packer, C., Gao, K., Kos, J., Kr{\"a}henb{\"u}hl, P., Koltun, V., and Song, D.
\newblock Assessing generalization in deep reinforcement learning.
\newblock \emph{arXiv preprint arXiv:1810.12282}, 2018.

\bibitem[Paszke et~al.(2019)Paszke, Gross, Massa, Lerer, Bradbury, Chanan,
  Killeen, Lin, Gimelshein, Antiga, Desmaison, Kopf, Yang, DeVito, Raison,
  Tejani, Chilamkurthy, Steiner, Fang, Bai, and Chintala]{pytorch}
Paszke, A., Gross, S., Massa, F., Lerer, A., Bradbury, J., Chanan, G., Killeen,
  T., Lin, Z., Gimelshein, N., Antiga, L., Desmaison, A., Kopf, A., Yang, E.,
  DeVito, Z., Raison, M., Tejani, A., Chilamkurthy, S., Steiner, B., Fang, L.,
  Bai, J., and Chintala, S.
\newblock Pytorch: An imperative style, high-performance deep learning library.
\newblock In Wallach, H., Larochelle, H., Beygelzimer, A., d\textquotesingle
  Alch\'{e}-Buc, F., Fox, E., and Garnett, R. (eds.), \emph{Advances in Neural
  Information Processing Systems 32}, pp.\  8024--8035. Curran Associates,
  Inc., 2019.

\bibitem[Pomerleau(1989)]{pomerleau1989alvinn}
Pomerleau, D.~A.
\newblock Alvinn: An autonomous land vehicle in a neural network.
\newblock In \emph{Advances in neural information processing systems}, pp.\
  305--313, 1989.

\bibitem[Ramachandran \& Amir(2007)Ramachandran and
  Amir]{ramachandran2007bayesian}
Ramachandran, D. and Amir, E.
\newblock Bayesian inverse reinforcement learning.
\newblock In \emph{IJCAI}, volume~7, pp.\  2586--2591, 2007.

\bibitem[Reddy et~al.(2019)Reddy, Dragan, and Levine]{reddy2019sqil}
Reddy, S., Dragan, A.~D., and Levine, S.
\newblock Sqil: Imitation learning via reinforcement learning with sparse
  rewards.
\newblock \emph{arXiv preprint arXiv:1905.11108}, 2019.

\bibitem[Ross et~al.(2011)Ross, Gordon, and Bagnell]{ross2011reduction}
Ross, S., Gordon, G., and Bagnell, D.
\newblock A reduction of imitation learning and structured prediction to
  no-regret online learning.
\newblock In \emph{Proceedings of the fourteenth international conference on
  artificial intelligence and statistics}, pp.\  627--635. JMLR Workshop and
  Conference Proceedings, 2011.

\bibitem[Schaal(1999)]{schaal1999imitation}
Schaal, S.
\newblock Is imitation learning the route to humanoid robots?
\newblock \emph{Trends in cognitive sciences}, 3\penalty0 (6):\penalty0
  233--242, 1999.

\bibitem[Schulman et~al.(2016)Schulman, Moritz, Levine, Jordan, and
  Abbeel]{gae}
Schulman, J., Moritz, P., Levine, S., Jordan, M., and Abbeel, P.
\newblock High-dimensional continuous control using generalized advantage
  estimation.
\newblock In \emph{International Conference on Learning Representations
  (ICLR)}, 2016.

\bibitem[Schulman et~al.(2017)Schulman, Wolski, Dhariwal, Radford, and
  Klimov]{schulman2017proximal}
Schulman, J., Wolski, F., Dhariwal, P., Radford, A., and Klimov, O.
\newblock Proximal policy optimization algorithms.
\newblock \emph{arXiv preprint arXiv:1707.06347}, 2017.

\bibitem[Shelhamer et~al.(2016)Shelhamer, Mahmoudieh, Argus, and
  Darrell]{shelhamer2016loss}
Shelhamer, E., Mahmoudieh, P., Argus, M., and Darrell, T.
\newblock Loss is its own reward: Self-supervision for reinforcement learning.
\newblock \emph{arXiv preprint arXiv:1612.07307}, 2016.

\bibitem[Shon et~al.(2006)Shon, Grochow, Hertzmann, and Rao]{shon2006learning}
Shon, A., Grochow, K., Hertzmann, A., and Rao, R.~P.
\newblock Learning shared latent structure for image synthesis and robotic
  imitation.
\newblock In \emph{Advances in neural information processing systems}, pp.\
  1233--1240, 2006.

\bibitem[Stadie et~al.(2017)Stadie, Abbeel, and Sutskever]{stadie2017third}
Stadie, B.~C., Abbeel, P., and Sutskever, I.
\newblock Third-person imitation learning.
\newblock \emph{arXiv preprint arXiv:1703.01703}, 2017.

\bibitem[Sun et~al.(2019)Sun, Zhan, Chan, and Tomizuka]{sun2019behavior}
Sun, L., Zhan, W., Chan, C.-Y., and Tomizuka, M.
\newblock Behavior planning of autonomous cars with social perception.
\newblock In \emph{2019 IEEE Intelligent Vehicles Symposium (IV)}, pp.\
  207--213. IEEE, 2019.

\bibitem[Torabi et~al.(2018)Torabi, Warnell, and Stone]{torabi2018behavioral}
Torabi, F., Warnell, G., and Stone, P.
\newblock Behavioral cloning from observation.
\newblock \emph{arXiv preprint arXiv:1805.01954}, 2018.

\bibitem[Weber et~al.(2017)Weber, Racani{\`e}re, Reichert, Buesing, Guez,
  Rezende, Badia, Vinyals, Heess, Li, et~al.]{weber2017imagination}
Weber, T., Racani{\`e}re, S., Reichert, D.~P., Buesing, L., Guez, A., Rezende,
  D.~J., Badia, A.~P., Vinyals, O., Heess, N., Li, Y., et~al.
\newblock Imagination-augmented agents for deep reinforcement learning.
\newblock \emph{arXiv preprint arXiv:1707.06203}, 2017.

\bibitem[Ziebart et~al.(2008)Ziebart, Maas, Bagnell, and
  Dey]{ziebart2008maximum}
Ziebart, B.~D., Maas, A.~L., Bagnell, J.~A., and Dey, A.~K.
\newblock Maximum entropy inverse reinforcement learning.
\newblock In \emph{Aaai}, volume~8, pp.\  1433--1438. Chicago, IL, USA, 2008.

\end{thebibliography}

\newpage
\cleardoublepage
\newpage

\section{Appendix}

\subsection{Auxiliary prediction loss}
\begin{figure}[h]
\centering
  \centering
  \includegraphics[width=0.8\linewidth]{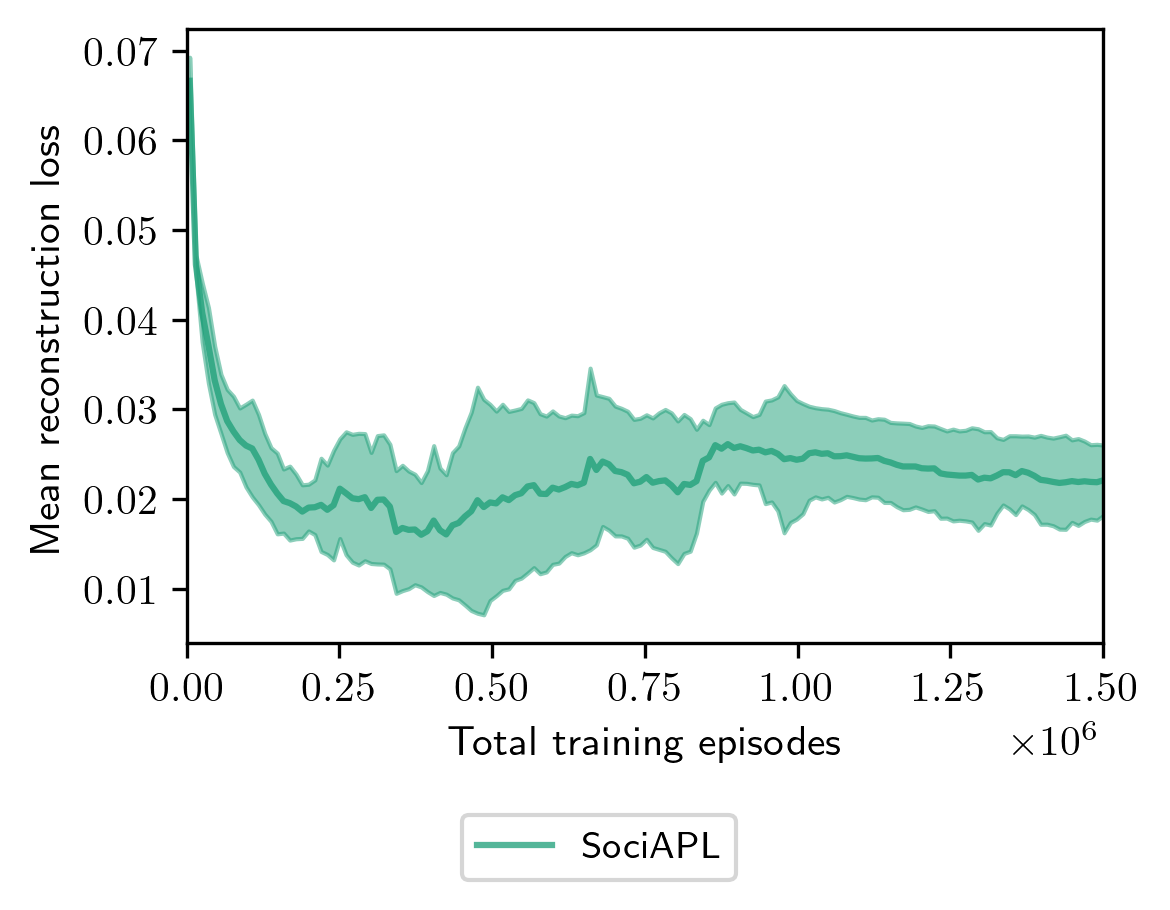}
  \caption{Next-state auxiliary prediction loss (in Mean Absolute Error (MAE) over the course of training the \textsc{social ppo + aux rec} agents shown in Figure \ref{fig:main_result}.}
  \label{fig:apl}
\end{figure}

Figure \ref{fig:apl} shows the training curve for the next-state auxiliary prediction loss of Equation \ref{eq:mae_next_state}, for the five agent seeds plotted in Figure \ref{fig:main_result} with auxiliary prediction losses. The curve shows that the agent is effectively learning to predict the next state with low mean absolute error. However, because the agent's policy is changing at the same time, the prediction problem is non-stationary, which means that the loss does not always decrease. If the agent discovers a new behavior, the model will be required to predict new state transitions not previously experienced. 

\subsection{Training without an exploration penalty}
\label{app:no_penalty}
\begin{figure}[h]
\centering
  \centering
  \includegraphics[width=0.8\linewidth]{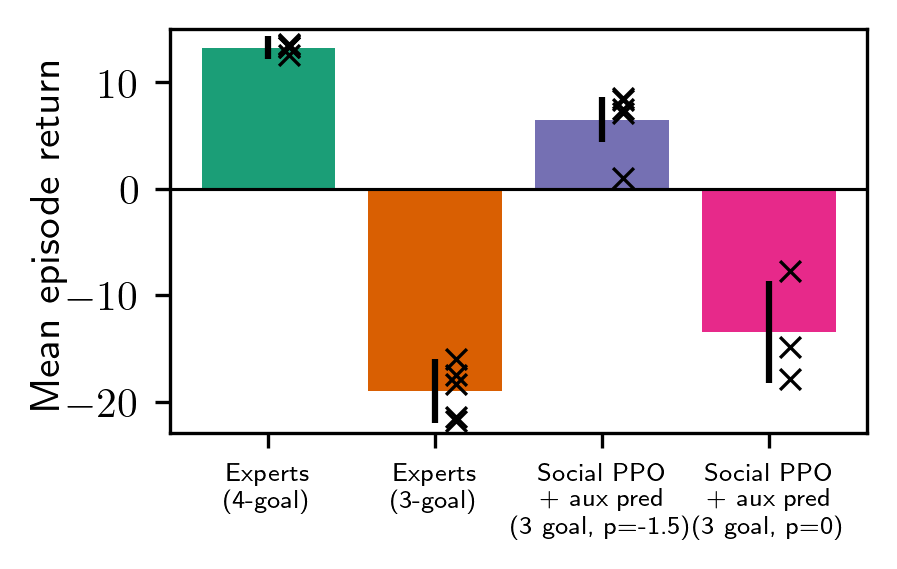}
  \caption{Next-state auxiliary prediction loss (in Mean Absolute Error (MAE) over the course of training the \textsc{social ppo + aux rec} agents shown in Figure \ref{fig:main_result}.}
  \label{fig:no_penalty}
\end{figure}

The Goal Cycle environments include a penalty for reaching the goals in the incorrect order. We conducted an additional experiment to investigate whether social learning emerges effectively without the exploration penalty. 
Figure \ref{fig:no_penalty} compares the transfer results of agents trained in Goal Cycle with the standard exploration penalty of $p=-1.5$, to agents trained with no penalty ($p=0$). Without the penalty, agents in Goal Cycle lack an incentive to learn from experts in the environment. The solitary policies they learn mirror those of the three-goal experts, reflected by similar performance in the four-goal transfer environment. This mirrors the finding that animals engage in social learning when individual exploration is most costly or dangerous \cite{Laland2004socialstrategies}.

\subsection{Identifying the experts}
\begin{figure}[h]
\centering
  \centering
  \includegraphics[width=.9\linewidth]{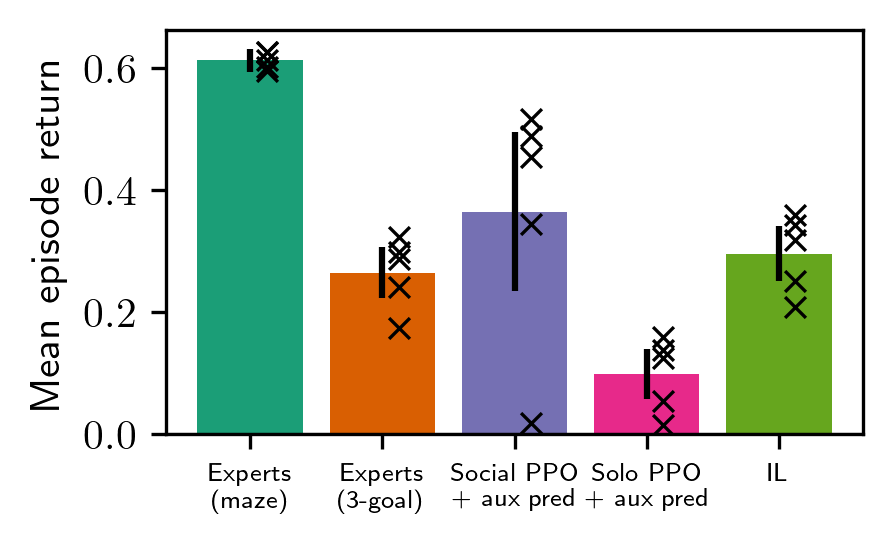}

  \caption{Distractors}
  \label{fig:distractors}
\end{figure}

In most of our experiments, we assume that social learners are trained in the presence of two high-performing expert agents. We conduct an additional experiment to determine if social learners can learn effectively in the presence of both high-performing experts, and non-experts. In this new experiment, the transfer task experts are accompanied by two non-expert distractor agents that take random movement actions. As shown by the maze transfer results in Figure \ref{fig:distractors}, our method is robust to the presence of these distractions and still outperforms the baselines.

\subsection{Imitation learning details}
\begin{figure}[h]
\centering
  \centering
  \includegraphics[width=0.8\linewidth]{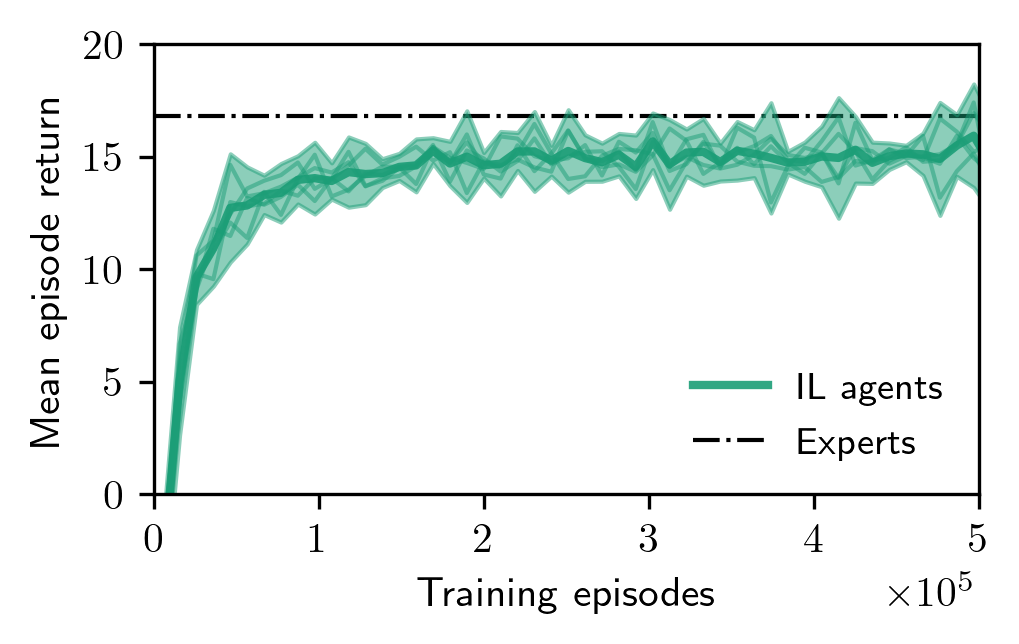}
  \caption{Returns over the course of training imitation learning agents in the 3 goal Goal Cycle environment with experts.}
  \label{fig:il_logp_training}
\end{figure}

The imitation learning agents shown in Section~\ref{sec:experiments} were trained with behavior cloning\cite{bain1995framework}. The model architecture is similar to that of the textsc{social ppo + aux rec} agents, but layers specific to the value function and next state prediction were unused. The policy parameters $\theta$ were optimized with Adam\cite{adam} to minimize the loss
$$
L_{\text{BC}} =  \mathbb{E_{D}} - \log \pi_{\theta}(a_{expert} | s_{demo}),
$$
where $D$ is a dataset of expert trajectories consisting of states $s_{demo}$ and expert actions $a_{expert}$. Figure~\ref{fig:il_logp_training} shows the returns of imitation learning agents during training.

\subsection{Q-learning learning in sparse reward environments}
\label{sec:0rewardtd}
Before a Temporal Difference (TD) learning agent has received any reward, it will be difficult for it to learn to model transition dynamics. 
Consider as an example deep Q-learning, in which the Q-function is parameterized by a neural network which encodes the state using a function $f_\theta(s)$. Assume the network is randomly initialized such that all Q-values are small, random values; \textit{i.e.} $\forall a \in \mathcal{A}, s \in \mathcal{S}, Q(a,s) = \epsilon \sim \mathcal{N}(0,0.1)$. 
Assume that the agent has not yet learned to navigate to the goal, and has received zero rewards so far during training.
Therefore, when the agent observes the experience $(s,a,r=0,s')$, the Q-learning objective is:
\begin{align}
    J(\theta) &= (r + \gamma \max_{a'}Q(s',a') - Q(s,a))^2 \\
    &= (0 + \gamma \epsilon_1 - \epsilon_2)^2
\end{align}
In effect, this induces a correlation between $Q(s',a')$ and $Q(s,a)$, and consequently $f_\theta(s')$ and $f_\theta(s)$, as a result of observing the state transition $s \rightarrow s'$. However, as the agent continues to receive zero reward, all Q-values will be driven toward zero. Once this occurs, even if the agent observes a useful novel state such as $\tilde{s}$, our equation becomes:
\begin{align}
    (r + \gamma \max_{a'}Q(s',a') - Q(\tilde{s},a))^2 = (0 - Q(\tilde{s},a))^2
\end{align}
such that the objective forces the value of $Q(\tilde{s},a)$ to be zero. In this case, modeling transitions into or out of state $\tilde{s}$ is not required in order to produce an output of zero, since all other Q-values are already zero.

\subsection{Environment details}
The environments used in this paper were originally based on Minigrid \citep{gym_minigrid}. The partial states constituting agent observations are 27$\times$27 RGB images corresponding to 7$\times$7 grid tiles. There are 7 possible actions, though  only three actions (rotate left, rotate right, move forward) are relevant in the experiments discussed in this paper. Agents are unable to see or move through the obstructive tiles that clutter their environments, and obstructed regions of their partial views appear as purple. However, agents can both see and move through goal tiles as well as other agents.

\begin{figure}[t]
\centering
  \centering
  \includegraphics[width=.8\linewidth]{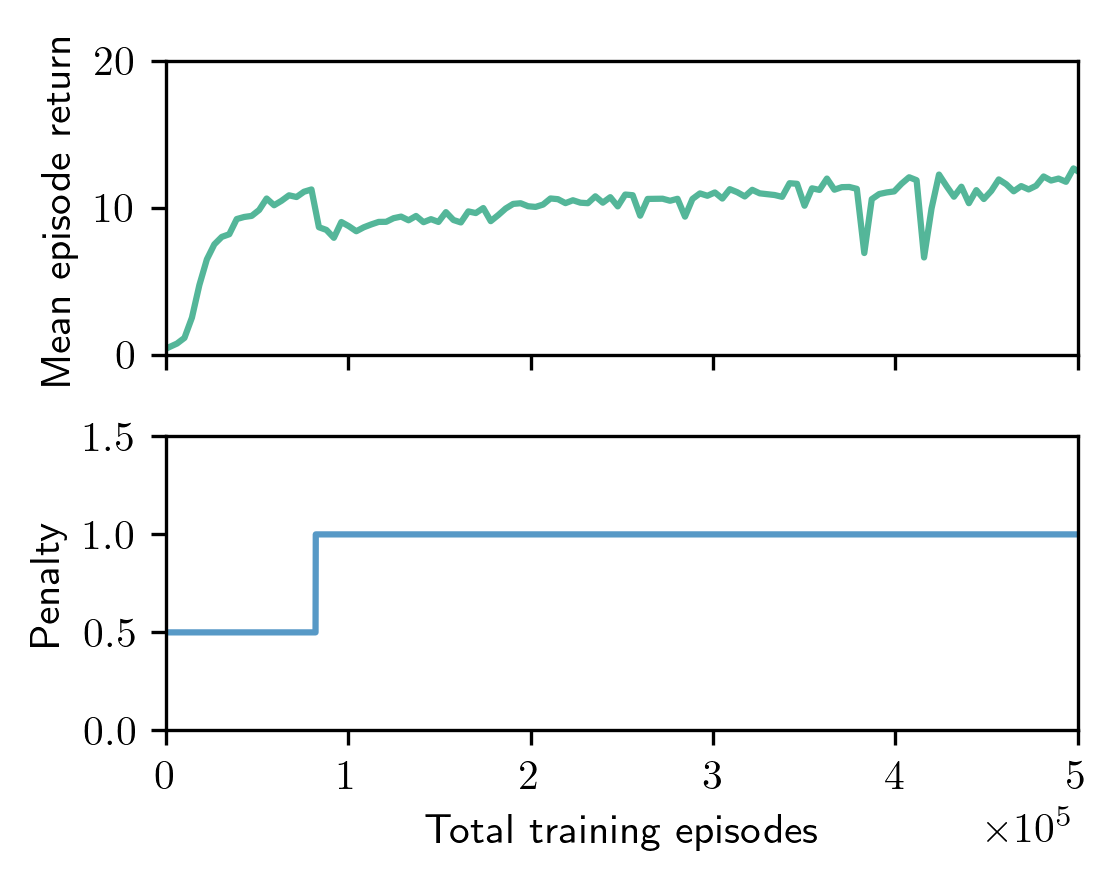}
  \caption{We use a penalty curriculum to obtain experts in environments where exploration is expensive. In the scenario visualized here, an agent trained for 81920 episodes in Goal Cycle environments with penalty of 0.5, then continued with a penalty of 1.0}
  \label{fig:penalty_curriculum}
\end{figure}

When the penalty for individual exploration in Goal Cycle environments is large, agents are unable to learn effective strategies. We used a penalty curriculum to obtain experts for such tasks as shown in Figure \ref{fig:penalty_curriculum}.

\begin{figure}[t]
\centering
  \centering
  \includegraphics[width=.8\linewidth]{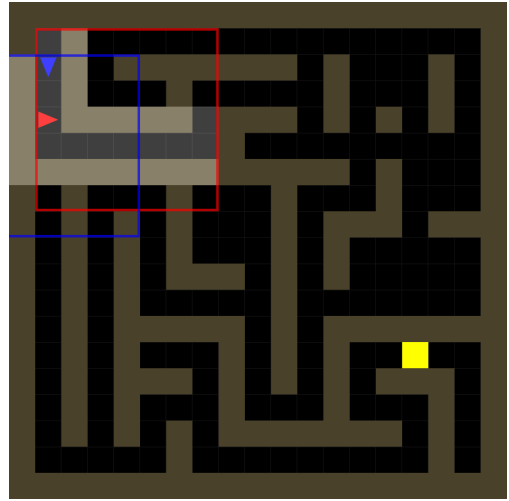}
  \caption{Sample 19x19 maze environment.}
  \label{fig:maze_environment}
\end{figure}

\subsection{Network architecture details}
The value, policy, and auxiliary task networks share three convolution layers, a fully connected layer, and an LSTM layer. Values and policies are computed with two fully connected layers, and the prediction-based auxiliary branch has a fully connected layer followed by transposed convolution layers that mirror the input convolution layers. The convolution and transposed convolution use leaky ReLU activation functions; all other layers use tanh activation functions.

\begin{itemize}
\item Shared input layers:
\begin{itemize}
    \item Conv $(3, 32)$ 3x3 filters, stride 3, padding 0,
    \item Conv $(32, 64)$ 3x3 filters, stride 1, padding 0,
    \item Conv $(64, 64)$ 3x3 filters, stride 1, padding 0,
    \item FC $(576, 192)$,
    \item LSTM $(192, 192)$
\end{itemize}
\item Value MLP:
\begin{itemize}
    \item FC $(192, 64)$,
    \item FC $(64, 64)$,
    \item FC $(64, 1)$
\end{itemize}
\item Policy MLP:
\begin{itemize}
    \item FC $(192, 64)$,
    \item FC $(64, 64)$,
    \item FC $(64, 7)$
\end{itemize}
\item Auxiliary state prediction layers:
\begin{itemize}
    \item FC $(192+7, 576)$,
    \item DeConv $(64, 64)$ 3x3 filters, stride 1, padding 0,
    \item DeConv $(64, 32)$ 3x3 filters, stride 1, padding 0,
    \item DeConv $(32, 3)$ 3x3 filters, stride 3, padding 0
\end{itemize}
\end{itemize}
Altogether there are 668555 parameters. We experimented with smaller (357291-parameter) networks but did not observe a significant performance difference. The networks were sized to roughly saturate the available (desktop) compute resources.

\subsection{Hyperparameters}

Each agent uses a single Adam optimizer \citep{adam} to update its parameters. Each of the novice agents was trained with a learning rate of $1\mathrm{e}-4$. The expert agents were trained with a learning rate of $1\mathrm{e}-5$.
Weights for all agents in the generalization experiments as well as the imperfect experts in textsc{social ppo + aux rec (imperfect experts)} were kept frozen and not updated.

Each parameter update consists of 20 sequential mini-batch updates with the same batch of rollouts (128 episodes). Each mini-batch consists of a uniform random sample of trajectories from that batch. Hidden states are stored alongside the trajectories, and the initial hidden state for each mini-batch trajectory is retrieved from these stored values. Hidden states and advantage values for the entire batch are re-calculated every 2 mini-batches. 
The mini-batch iteration ceases if $\textsc{KL}(\pi, \pi^{rollout})$ exceeds a target of 0.01. If for any mini-batch the estimated divergence exceeds a hard limit of 0.03, the update terminates and all changes to the network parameters and optimizer state are reverted.

\begin{table}[h]
\begin{tabular}{|l|l|}
\hline
batch size & 128 episodes \\ \hline
mini-batches per batch & 20 \\ \hline
mini-batch num trajectories & 512 \\ \hline
mini-batch trajectory length & 16 \\ \hline
hidden state/advantage update interval & 2 minibatches \\ \hline
return discount $\gamma$ & 0.993 \\ \hline
GAE-$\lambda$ & 0.97 \\ \hline
PPO clip ratio & 0.2 \\ \hline
KL target & 0.01 \\ \hline
KL hard limit & 0.03 \\ \hline
\end{tabular}
\end{table}

For each mini-batch iteration, the loss used to update agent parameters is $$L^{\text{total}} = L^{\pi}(\theta) + c_{V}\cdot L^{V}(\theta) + c_{\text{aux}}\cdot L^{\text{aux}}(\theta) - c_{\text{ent}}\cdot L^{\text{ent}}(\theta),$$ where the policy loss $L^{\pi}$ is computed with PPO-clip \citep{schulman2017proximal}  and GAE \citep{gae}, the value loss $L^{V}$ is the mean squared error of the values estimated for each step in the trajectory, $L^{\text{ent}}(\theta)$ is the policy entropy bonus, and $L^{\text{aux}}(\theta)$ is the auxiliary prediction loss (Equation ~\ref{eq:mae_next_state}). The loss scaling coefficients used in our experiments are $c_{V} = 0.1$, $c_{\text{ent}} = 1\mathrm{e}-5$, and $c_{\text{aux}} = 3$. 

The prestige decay constant $\alpha_c$ used for the Goal Cycle environments (i.e. Equation \ref{eq:prestige_equation}) was 0.99.

In general we sought hyperparameters that enable stable training. We experimented with mini-batch sizes varying from 32 to 1025 trajectories and found training to be more stable with larger mini-batches. Training was less stable with learning rates higher than $1\mathrm{e}-4$. 

We randomized seeds for both the network parameter initialization and environment generation for each trial of each experiment.  

\subsection{Compute}

The experiments in this paper were performed primarily on a desktop computer with an AMD Ryzen 3950x CPU and two Nvidia GTX 1080TI GPUs, as well as g4dn.8xlarge instances provisioned on Amazon AWS. Either system can run two or three trials simultaneously, each consisting of three agents training together in a shared environment. Collecting experience and updating parameters were comparably time consuming, and a single 1.5M episode (375M step) 3-agent training run took about 30 hours.
We used Ubuntu 18.04 with python3.8 and all neural networks are implemented in PyTorch v1.6 \citep{pytorch}. Training metrics were logged with Weights and Biases \citep{wandb}.

\end{document}